\documentclass{article}

\usepackage[final,main]{neurips_2026}

\usepackage[utf8]{inputenc} 
\usepackage[T1]{fontenc}    
\usepackage{hyperref}       
\usepackage{url}            
\usepackage{booktabs}       
\usepackage{amsfonts}       
\usepackage{nicefrac}       
\usepackage{microtype}      
\usepackage{xcolor}         

\usepackage{amsmath,amssymb}
\DeclareMathOperator*{\argmin}{arg\,min}
\usepackage{graphicx}
\usepackage{subcaption}
\usepackage{multirow}
\usepackage{makecell}

\usepackage{tikz}
\usepackage{caption}
\usepackage{rotating}
\usepackage{placeins}
\usepackage{wrapfig}

\newcommand{\shin}[1]{\textcolor{black}{#1}}

\title{Slower Generalization, Faster Memorization:\\
A Sweet Spot in Algorithmic Learning}

\author{%
  Shin So \qquad
  Kyelim Lee \qquad
  Albert No\thanks{Correspondence to: \texttt{albertno@yonsei.ac.kr}}\\[0.6em]
  Department of Artificial Intelligence, Yonsei University
}

\begin{document}
\maketitle

\setlength{\textfloatsep}{10pt plus 2pt minus 2pt}
\setlength{\intextsep}{10pt plus 2pt minus 2pt}

\setcounter{topnumber}{1}
\setcounter{bottomnumber}{0}
\setcounter{totalnumber}{1}
\makeatletter
\setlength{\@fptop}{0pt}
\makeatother

\begin{abstract}
Once generalization becomes possible, does adding more data always make learning faster?
In Needleman–Wunsch matrix generation, we find that small Transformers reach high exact-match validation accuracy
in the fewest optimizer updates at an intermediate dataset size, while larger datasets converge more slowly.
A random-suffix probe shows that, at larger dataset sizes, models fit the structured matrices before memorizing their random suffixes,
suggesting that shared rule learning aids training-set fitting.
To examine what makes rules easier to learn, we study elementary cellular automata (ECA),
in which each binary cell updates from its own state and those of its two neighbors.
These experiments identify lower-order predictive signals--output biases or predictive relationships involving only a few input bits--as a key factor in learnability.
In Rule 90, finite-sample fluctuations create output biases and single-input dependencies absent from the uniform population.
These signals weaken as the dataset grows.
Biased training that introduces these signals accelerates generalization on an unchanged uniform validation distribution.
These results support an explanation for faster convergence at intermediate dataset sizes:
sampling fluctuations can facilitate rule acquisition, and larger datasets weaken this advantage.
\end{abstract}

\section{Introduction} \label{sec:intro}

Once a model has enough examples to generalize, can more examples make learning slower?
Empirical scaling laws establish the value of additional data for predictive performance when model size and compute are scaled appropriately~\citep{hestness2017deep,kaplan2020scaling,hoffmann2022training}.
In algorithmic learning, studies of grokking and critical dataset sizes further show that more data can enable generalization and shorten the delay before it emerges~\citep{power2022grokking,liu2022towards,zhu2024critical}.
These findings make continued acceleration a natural expectation.
We examine how far that expectation extends,
distinguishing the amount of data that makes generalization possible from the amount that makes it fastest.

\begin{figure}[tp]
    \centering
    \includegraphics[width=\linewidth]{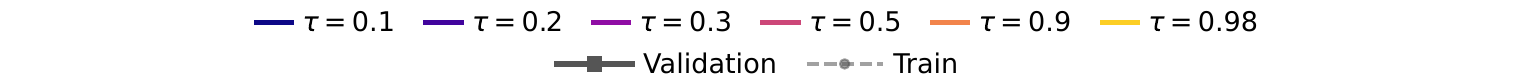}\par\vspace{2pt}
    \begin{subfigure}[t]{0.49\linewidth}
        \centering
        \includegraphics[width=\linewidth]{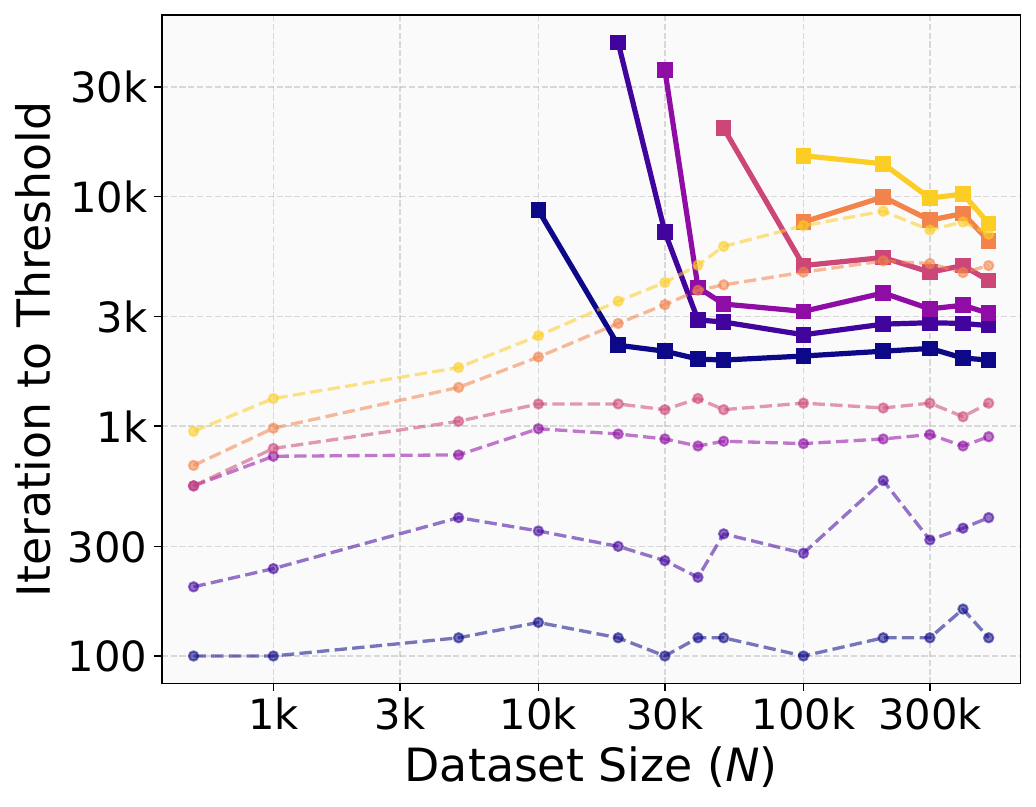}
        \caption{Three-digit multiplication.}
        \label{fig:figure1a}
    \end{subfigure}
    \hfill
    \begin{subfigure}[t]{0.49\linewidth}
        \centering
        \includegraphics[width=\linewidth]{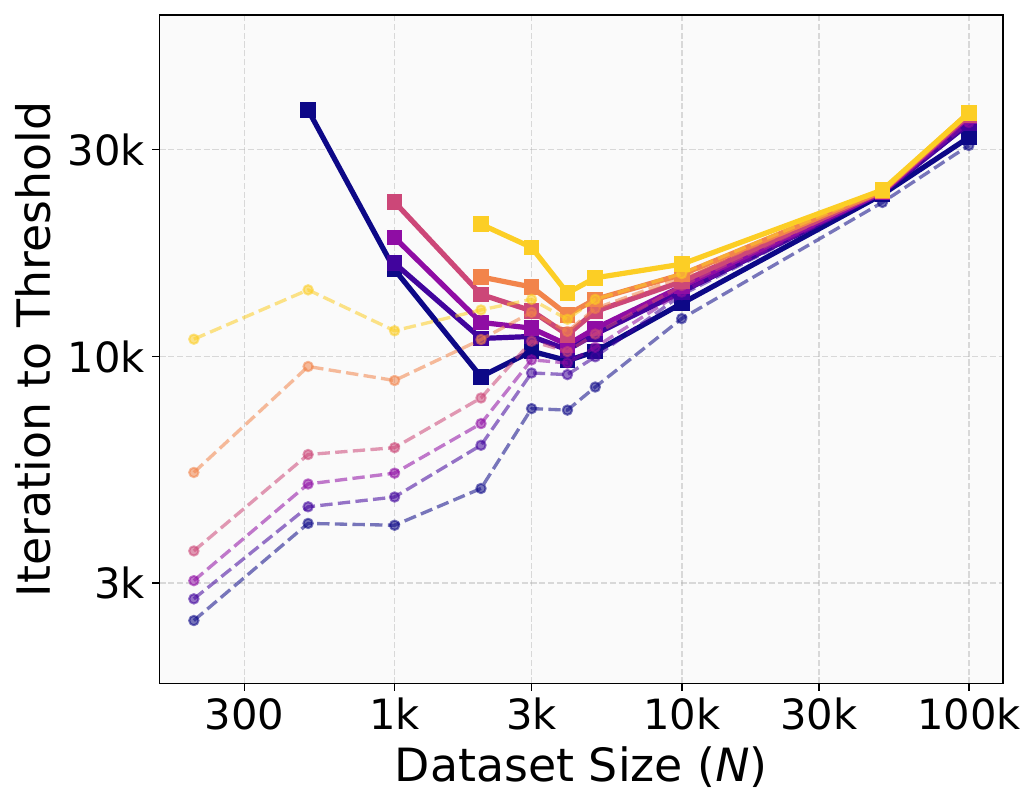}
        \caption{NW matrix generation, $L=5$, $D=6$.}
        \label{fig:figure1b}
    \end{subfigure}
    \caption{Dataset-size sweeps for multiplication and NW matrix generation. Each curve reports the mean recorded number of optimizer updates needed to reach exact-match threshold \(\tau\); lower values mean faster convergence. Colors denote thresholds \(\tau\), solid lines show validation crossings \(I_V(\tau;N)\), and dashed lines show training crossings \(I_T(\tau;N)\). Multiplication shows no pronounced post-threshold slowdown. NW reaches validation thresholds fastest at intermediate sizes and more slowly at the largest dataset size in the sweep.}
    \label{fig:figure1}
\end{figure}

We study this question in Needleman–Wunsch (NW) matrix generation,
a task derived from sequence alignment~\citep{needleman1970general}.
Given two input strings, a small decoder-only Transformer learns to generate the score matrix defined by the NW dynamic program.
A prediction counts as correct only when every matrix entry matches the target.
We find that validation accuracy reaches a given threshold fastest at an intermediate dataset size (Figure~\ref{fig:figure1}).
Beyond the optimal dataset size, larger datasets require more optimizer updates to reach the same threshold.
We call this increase in convergence time among dataset sizes that already permit generalization the large-data slowdown.

The contrast with multiplication makes this finding particularly striking.
Arithmetic tasks are widely used to study algorithmic learning and grokking~\citep{power2022grokking,liu2022towards,lee2023teaching}, making three-digit multiplication a natural comparison.
In our baseline, which predicts the product directly, validation convergence becomes faster and then roughly plateaus as the dataset grows,
with no comparable slowdown over the tested range (Figure~\ref{fig:figure1}).
Thus, additional examples of a deterministic rule can have qualitatively different effects on learning speed across tasks.

The slowdown extends the grokking picture of how training-set fitting and generalization unfold.
Increasing dataset size can shorten the delay between fitting and generalization~\citep{power2022grokking,liu2022towards};
NW exhibits a further regime in which this delay has narrowed,
yet additional data slows convergence on both sets (Section~\ref{sec:main-results}).
Near the onset of weak generalization, we also find that adding examples can accelerate training-set fitting,
suggesting that partial rule learning already helps explain training predictions.
To examine this interaction, we append a fixed, randomly assigned four-bit suffix to each NW target as a reference for example-specific memorization (Section~\ref{sec:mechanism}).
As generalization emerges, matrix training accuracy advances ahead of suffix accuracy,
and their early gap approximately tracks validation accuracy.
These observations support a contribution from partial rule learning to training-set fitting.
Meanwhile, memorizing the suffix becomes slower as the dataset grows,
revealing an increasing cost of fitting example-specific associations.

To investigate why the slowdown depends on the task,
we study elementary cellular automata (ECA),
in which each binary cell evolves according to a local rule based on its own state and those of its two neighbors (Section~\ref{sec:eca}).
We identify lower-order predictive signals—output biases or predictive relationships involving only a few input bits—as a key factor in learnability.
For Rule 90, output biases and single-input predictive relationships are absent under the uniform population distribution but arise from finite-sample deviations in the training data.
These signals weaken as the dataset grows.
Biased training that introduces these signals accelerates generalization on an unchanged uniform validation distribution, while matched datasets balanced to eliminate first-transition single-input dependencies take longer to fit.
These results support a mechanism by which sampling fluctuations can facilitate rule acquisition at intermediate dataset sizes.
Together, the NW sweeps, random-suffix probe, and ECA experiments suggest complementary roles for the cost of example-specific fitting and the weakening of empirical predictive signals in the large-data slowdown.

\section{Related Work} \label{sec:related}
\paragraph{Grokking and train--validation timing.} Grokking is delayed generalization, in which training accuracy becomes high long before validation accuracy rises on small algorithmic datasets~\citep{power2022grokking}. Subsequent work connects the delay to structured representations~\citep{liu2022towards}, circuit formation and efficiency~\citep{nanda2023progress,varma2023explaining}, implicit-bias transitions~\citep{lyu2023dichotomy}, and internal subnetworks~\citep{minegishi2025bridging}. \citet{xu2026togrok} derive quantitative bounds on grokking time in ridge regression as a function of training hyperparameters. Dataset-size sweeps in earlier work also show that more data can shorten the delay before generalization~\citep{power2022grokking,liu2022towards}. Our work shares the focus on training--validation timing but asks whether faster generalization continues as the dataset grows.

\paragraph{Dataset size, scaling, and critical thresholds.} Empirical scaling laws study how loss changes with data, parameters, and compute~\citep{hestness2017deep,kaplan2020scaling,hoffmann2022training}, while double descent shows that more data or training can sometimes hurt in certain effective-complexity regimes~\citep{nakkiran2021deep}. Critical-data-size work asks when generalization first becomes possible and often finds that additional data beyond this onset accelerates validation convergence~\citep{power2022grokking,liu2022towards,zhu2024critical}. Our question concerns the fastest convergence among dataset sizes that already permit generalization. Larger datasets can also be less effective under a fixed training budget in other algorithmic settings.
\citet{charton2024repeated} show benefits from repeating a smaller set of examples, \citet{wen2025sparse} report parity settings in which smaller datasets generalize while a larger dataset fails to learn within the budget,
and \citet{liu2026lessdata} connect sampling biases to relative layer growth and report parity speedups that persist after removing input biases and single-input label dependencies.
Our NW fitting diagnostics and ECA signal interventions examine a complementary, task-dependent aspect of this phenomenon.

\paragraph{Memorization and rule-mediated fitting.} Neural networks can memorize random labels~\citep{zhang2017rethinking}, yet often learn simple shared patterns before fitting noise~\citep{arpit2017closer}. Memorization can also be useful for high-accuracy learning when simple shared patterns are insufficient~\citep{feldman2020neural}; in language models, memorization has been studied through unintended memorization, extraction, and capacity estimates~\citep{carlini2019secret,carlini2021extracting,morris2025much}. Mechanistic accounts of grokking suggest that memorizing and generalizing solutions can differ in efficiency~\citep{nanda2023progress,varma2023explaining,liu2023omnigrok,huang2024unified}. NW matrix generation lets us observe this interaction through dataset-size sweeps.

\paragraph{Algorithmic and structured-output tasks.}
Controlled algorithmic tasks are useful because exact generalization and memorization can be measured cleanly. Prior work has used modular arithmetic, parity, arithmetic, and synthetic symbolic tasks to study delayed generalization and hidden progress \citep{power2022grokking,barak2022hidden,lee2023teaching,charton2024learning}. NW matrix generation differs from scalar arithmetic tasks because a short input pair induces a full dynamic-programming table whose cells are mutually constrained. This places NW among structured dynamic-programming problems such as edit distance, local sequence alignment, Viterbi decoding, and context-free parsing \citep{wagner1974string,smith1981identification,viterbi1967error,younger1967recognition}. We use NW as the primary evidence task and multiplication as a contrast baseline. ECA provides a controlled comparison of task learnability at fixed output length and recurrence structure.

\section{Dataset Size and the Speed of Algorithmic Learning} \label{sec:main-results}
We examine how training dataset size affects the number of optimizer steps needed to reach a given validation accuracy.
For each dataset size \(N\), we record the first evaluated checkpoint at which validation exact-match accuracy reaches a specified threshold, keeping the model and training protocol fixed within each sweep.
We also track the steps needed to reach training-accuracy thresholds to compare the timing of training-set fitting and generalization.
Figure~\ref{fig:figure1} summarizes these measurements for NW and multiplication, with lower values indicating faster convergence.

\subsection{Task Setup: NW Matrix Generation and Multiplication} \label{sec:tasks-models}
In Needleman--Wunsch (NW) matrix generation, a model predicts a sequence-alignment score matrix from two input strings~\citep{needleman1970general}. Let $\Sigma=\{A,C,G,T\}$ and let $x,y\in\Sigma^L$. Each input pair has a target matrix $M(x,y)\in\mathbb{Z}^{(L+1)\times(L+1)}$. Boundary entries use a fixed gap penalty, and interior entries obey
\begin{equation*}
    M_{i,j}=\max\{M_{i-1,j-1}+s(x_i,y_j),\; M_{i-1,j}+g,\; M_{i,j-1}+g\},
    \label{eq:nw}
\end{equation*}
where $s$ is the match/mismatch score and $g$ is the gap penalty. Unless otherwise stated, we use match score $5$, mismatch score $-4$, and gap penalty $-5$. The main experiments use $L=5$, so each input has ten symbols and the target is a $6\times 6$ matrix. The model predicts the serialized matrix autoregressively, and exact-match accuracy counts an example as correct only if every matrix entry is correct.
We use three-digit multiplication as a comparison task. Each input is a pair of three-digit integers and the target is the exact decimal product.

\begin{figure}[tp]
    \centering
    \includegraphics[width=\linewidth]{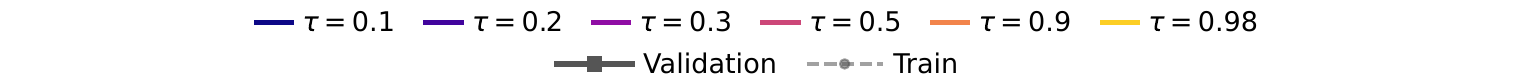}\par\vspace{2pt}
    \begin{subfigure}[t]{0.49\linewidth}
        \centering
        \includegraphics[width=\linewidth]{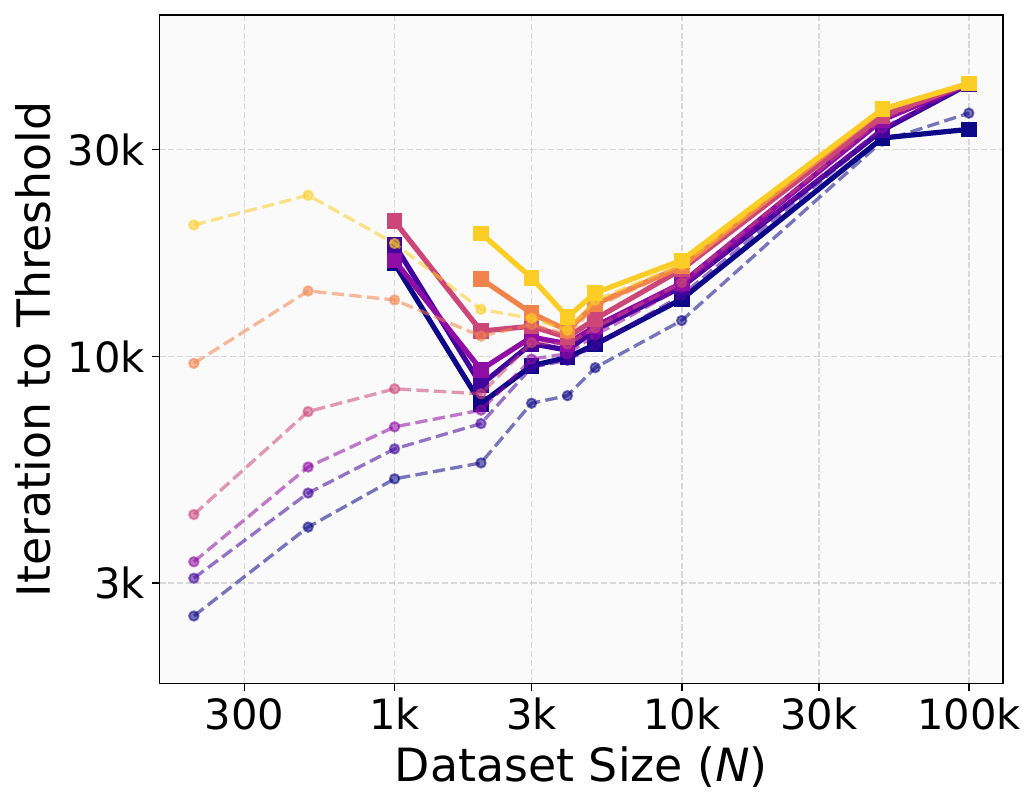}
        \caption{Depth 4.}
        \label{fig:figure2a}
    \end{subfigure}
    \hfill
    \begin{subfigure}[t]{0.49\linewidth}
        \centering
        \includegraphics[width=\linewidth]{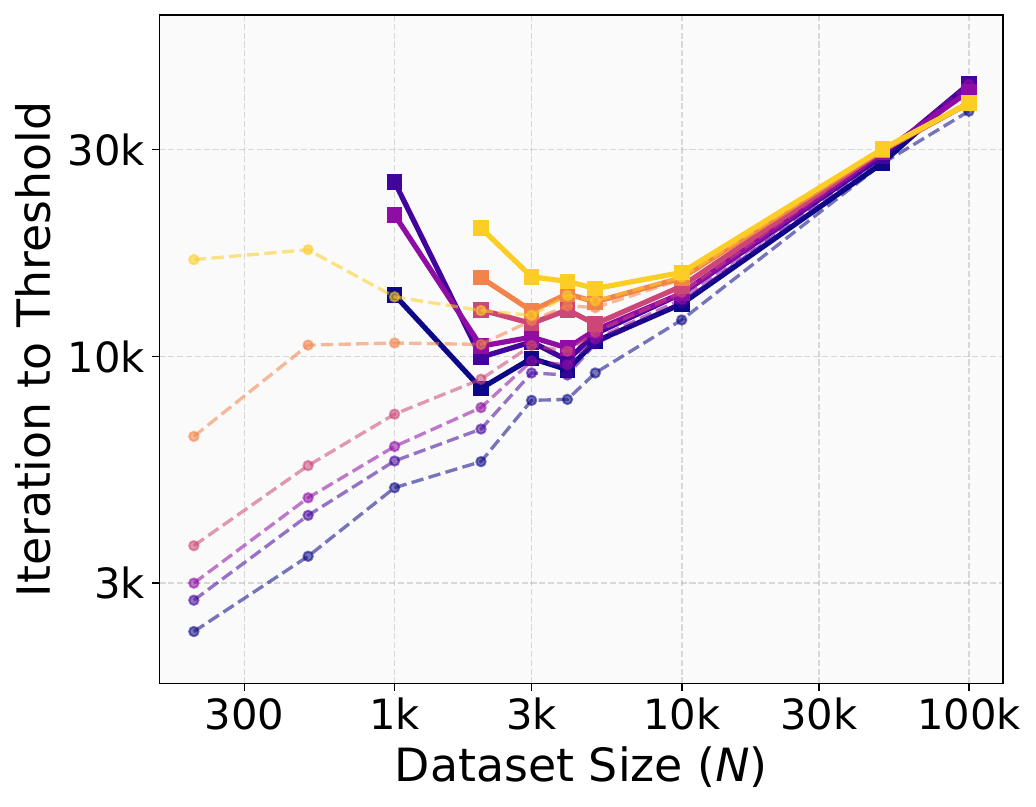}
        \caption{Depth 5.}
        \label{fig:figure2b}
    \end{subfigure}
    \caption{
    NW multi-threshold sweeps across depths \(D\in\{4,5\}\) at \(L=5\). Colors denote exact-match thresholds, solid lines show validation crossings, and dashed lines show training crossings. Across depths, validation crossing time is non-monotonic in dataset size, giving an intermediate-data sweet spot. In the same dataset-size band where weak validation thresholds first become reachable, training crossings can also become faster as \(N\) increases, supporting the partial-rule interpretation.
    }
    \label{fig:figure2}
\end{figure}

Both tasks use decoder-only Transformers with $D$ layers, $D$ attention heads, embedding width $64D$, and feed-forward width $256D$. Figure~\ref{fig:figure1} uses $D=6$, and Figure~\ref{fig:figure2} checks depths four and five. AdamW and the training schedule are fixed within each sweep. Additional details are given in Appendix~\ref{app:task_details}, with model and training settings in Appendix~\ref{app:training_details}.

\subsection{Crossing Times, Critical Size, and Optimal Size} \label{sec:crossing-times}

Let $\mathcal{D}_N$ denote a training set of $N$ examples, and let $A_T(t;N)$ and $A_V(t;N)$ denote the measured training and validation exact-match accuracies after $t$ optimizer updates. Each run has a budget of $T_{\mathrm{budget}}=50{,}000$ updates. For an individual run, let $\mathcal{T}_{\mathrm{eval}}$ denote the evaluated update counts within this budget. For threshold $\tau\in(0,1)$, the crossing times are the first evaluated checkpoints at which accuracy reaches at least $\tau$:
\begin{align*}
    I_T(\tau;N)&=\min\{t\in\mathcal{T}_{\mathrm{eval}}:A_T(t;N)\geq\tau\},\\
    I_V(\tau;N)&=\min\{t\in\mathcal{T}_{\mathrm{eval}}:A_V(t;N)\geq\tau\}.
\end{align*}
A run is censored when the threshold is not reached within the budget, and no finite crossing time is assigned.
A crossing records first attainment, without requiring the accuracy to remain above the threshold at subsequent checkpoints.
We use $\tau=0.98$ as the strict threshold and lower thresholds to detect partial competence.
Appendix~\ref{app:measurement-scope} describes the treatment of non-crossing runs.

For the NW dataset-size summaries, let $\overline I_V(\tau;N)$ denote the mean validation crossing time over runs that reach the threshold within the budget, and let $\mathcal{N}_\tau$ contain the tested sizes with at least one recorded validation crossing. The \emph{critical dataset size} is the smallest such tested size,
\[
    N_c(\tau)=\min\mathcal{N}_\tau.
\]
An \emph{optimal dataset size} minimizes the mean recorded validation crossing time,
\[
    N^\star(\tau)\in\argmin_{N\in\mathcal{N}_\tau}\overline I_V(\tau;N).
\]
Non-crossing runs are excluded from the mean; a dataset size is excluded only if it has no recorded crossing. These empirical summaries are conditional on threshold attainment and depend on the threshold, training budget, and tested sizes. A \emph{dataset-size sweet spot} exists when an optimal size lies strictly between the smallest and largest tested sizes.

\subsection{Multiplication versus NW} \label{sec:task-comparison}
Figure~\ref{fig:figure1} compares training and validation crossing times across dataset sizes and accuracy thresholds.
In multiplication, validation crossings generally become earlier as more examples are added, then remain stable across larger datasets.
Additional data therefore initially accelerates generalization, after which the gains largely saturate.
At lower accuracy thresholds, training crossings remain substantially earlier than the corresponding validation crossings.
This separation narrows at higher thresholds, but the validation curves show no pronounced upward turn at the large-data end of the sweep.

NW exhibits a different response. As dataset size increases from the smaller-data range,
training and validation crossing times move closer together.
At larger sizes, the two families of curves become closely aligned and both turn upward:
reaching the same accuracy on either split requires more optimizer updates.
Thus, training and validation thresholds can be reached close together in time while both take longer to attain as the dataset grows.
Unlike the approximate plateau observed in multiplication, 
NW validation crossing times attain their minima at intermediate dataset sizes.

\subsection{Two Findings: Faster Fitting and Slower Generalization} \label{sec:two-findings}
The NW curves motivate two complementary questions about the effects of additional data.
Near the onset of weak generalization, can adding examples make the training set faster to fit?
Among larger datasets that already permit generalization, does more data continue to reduce validation crossing time?
These questions concern different dataset-size ranges.
We examine the strict training threshold alongside lower validation thresholds to distinguish nearly complete training-set fitting from emerging validation competence.

\paragraph{Finding 1: more data can accelerate training-set fitting.}
In the dataset-size range where lower validation thresholds first become reachable, 
adding examples can reduce \(I_T(0.98;N)\), allowing the model to reach nearly perfect training accuracy in fewer updates despite the larger training set. 
Figure~\ref{fig:figure2} shows this behavior across model depths. 
Here, faster memorization refers to earlier completion of training-set fitting, not necessarily faster memorization of individual examples. 
The coincidence with emerging validation competence suggests that partial rule learning helps fit the training set. Even an incomplete rule can explain predictions across multiple training examples, 
reducing what remains to be memorized individually.

\paragraph{Finding 2: generalization is fastest at an intermediate dataset size.}
The NW validation curves exhibit a dataset-size sweet spot: the optimal dataset size \(N^\star(\tau)\) lies strictly between the smallest and largest tested sizes.
Beyond \(N^\star(\tau)\), larger datasets with recorded validation crossings require more optimizer updates on average to reach the same threshold, i.e., more data slows generalization.

To better understand these findings, Section~\ref{sec:mechanism} uses a random-suffix probe to examine how shared rule learning and example-specific fitting contribute to training progress.
Section~\ref{sec:eca} then investigates the large-data slowdown through the lens of task learnability.

\section{Rule Learning and the Cost of Example-Specific Fitting} \label{sec:mechanism}
Partial rule learning can accelerate fitting, yet larger NW datasets still generalize more slowly beyond the optimal dataset size.
If a shared rule makes training examples easier to fit, why does learning from more examples of the same rule require so many additional updates? The benefit of rule learning may be offset by a growing burden of example-specific fitting.
We examine that burden with a random-suffix probe, attaching a fixed, randomly assigned suffix to each NW target.

\subsection{Shared Rules Make Training Examples Easier to Fit} \label{sec:random-suffix}

By \emph{residual exact fitting}, we mean fitting training predictions that the model's currently learned rule does not yet explain. Those predictions may be resolved by learning more of the shared rule or through example-specific fitting, meaning memorization of individual input--target associations. A random suffix provides a reference for the latter because its assignments cannot be predicted from the NW rule, rather than a calibrated measure of the residual fitting cost of the original NW matrix.

We append a four-bit random binary suffix to each NW target. Each suffix is sampled once and held fixed throughout training.
The matrix follows the shared NW rule, whereas suffix assignments are independent across examples and must be memorized for the training inputs.
The NW rule cannot predict a new example's suffix.
The short suffix provides an unstructured comparison with the longer rule-generated matrix. 
Let \(A_V(t)\) be validation accuracy on the NW matrix, \(A_T(t)\) training accuracy on the NW matrix, and \(A_R(t)\) training accuracy on the random suffix.

\begin{figure}[tp]
    \centering
    \includegraphics[width=\linewidth]{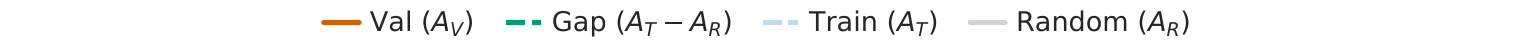}\par\vspace{2pt}
    \begin{subfigure}[t]{0.34\linewidth}
        \centering
        \includegraphics[width=\linewidth]{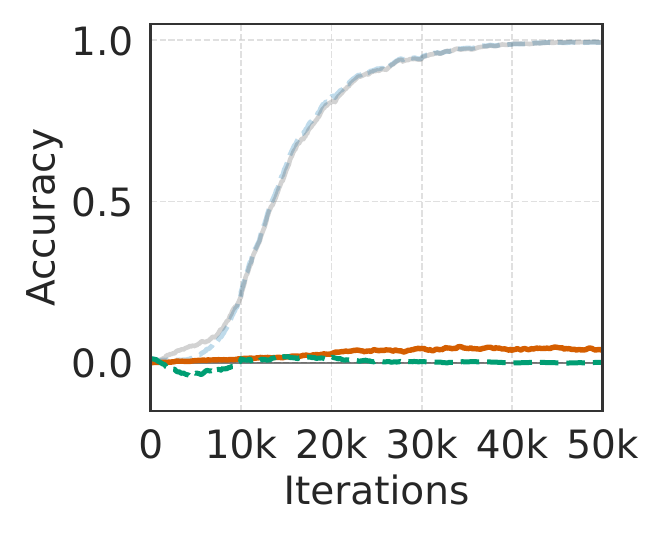}
        \caption{$N=1$k.}
        \label{fig:figure3a}
    \end{subfigure}%
    \hspace{6pt}%
    \begin{subfigure}[t]{0.28\linewidth}
        \centering
        \includegraphics[width=\linewidth]{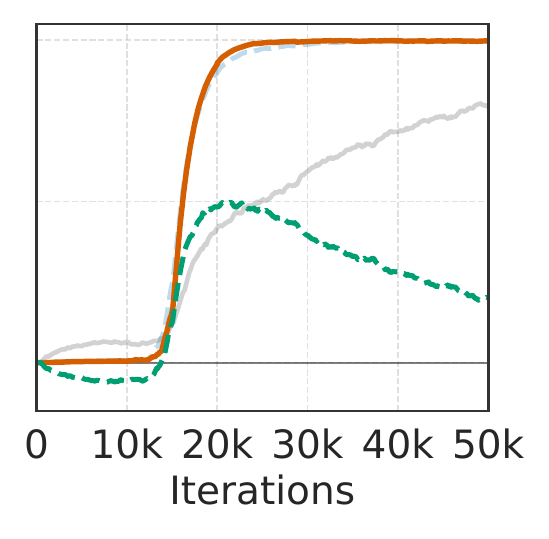}
        \caption{$N=10$k.}
        \label{fig:figure3b}
    \end{subfigure}%
    \hspace{6pt}%
    \begin{subfigure}[t]{0.28\linewidth}
        \centering
        \includegraphics[width=\linewidth]{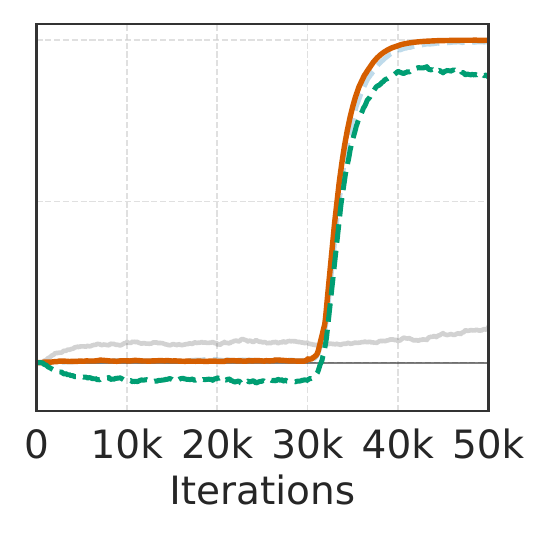}
        \caption{$N=100$k.}
        \label{fig:figure3c}
    \end{subfigure}
    \par\vspace{4pt}
    \begin{subfigure}[t]{0.34\linewidth}
        \centering
        \includegraphics[width=\linewidth]{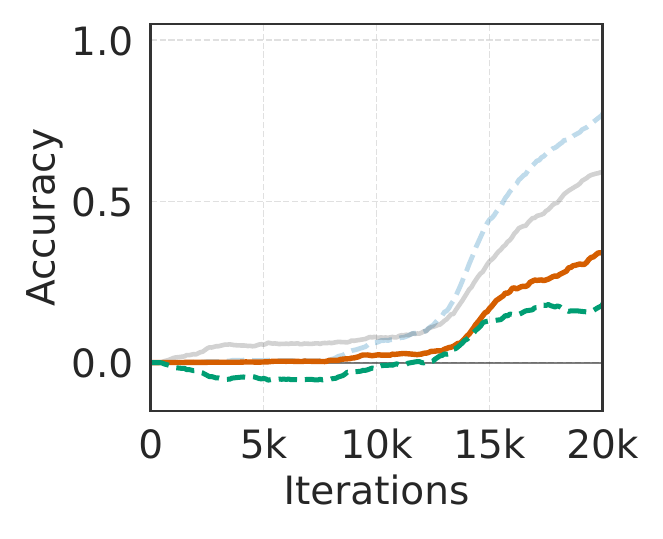}
        \caption{$N=4$k.}
        \label{fig:figure3d}
    \end{subfigure}%
    \hspace{6pt}%
    \begin{subfigure}[t]{0.28\linewidth}
        \centering
        \includegraphics[width=\linewidth]{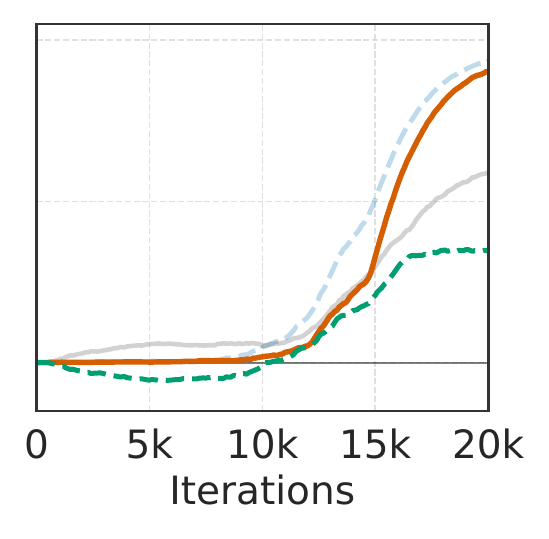}
        \caption{$N=5$k.}
        \label{fig:figure3e}
    \end{subfigure}%
    \hspace{6pt}%
    \begin{subfigure}[t]{0.28\linewidth}
        \centering
        \includegraphics[width=\linewidth]{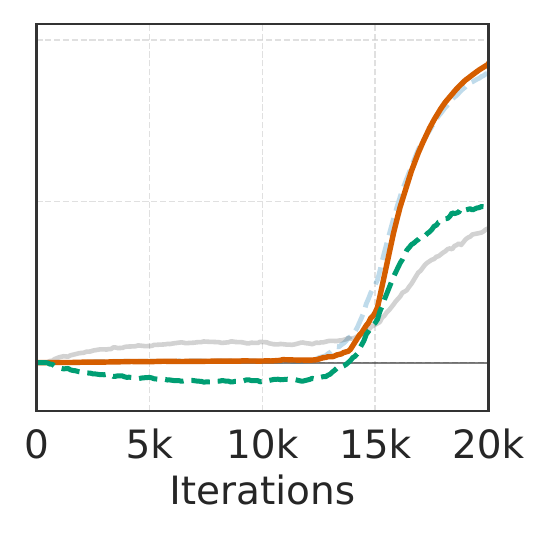}
        \caption{$N=10$k.}
        \label{fig:figure3f}
    \end{subfigure}
    \caption{Random-suffix probe and train--random gap. Each NW target includes a four-bit random suffix. Faint curves show training accuracy on the NW matrix \(A_T\) and random suffix \(A_R\). Dark curves show NW validation accuracy \(A_V\) and \(G=A_T-A_R\). \textbf{Top (a--c)}: full trajectories. As dataset size grows, the structured matrix is fit before the shorter random suffix. \textbf{Bottom (d--f)}: enlarged early-training views. Validation initially rises alongside the gap, suggesting that rule-mediated progress contributes to both training and validation accuracy. The approximate agreement is local to acquisition onset, not the entire trajectory.}
    \label{fig:figure3}
\end{figure}

Figure~\ref{fig:figure3}(a--c) compares these component accuracies. At small \(N\), matrix accuracy and suffix accuracy rise on similar timescales while validation remains low.
At larger \(N\), the model fits the rule-generated matrix before the much shorter random suffix.
This pattern is consistent with shared rule learning helping fit the matrix, while \(A_R\) tracks progress on the separately assigned random associations.

\subsection{Early Fitting Combines Rule Learning and Memorization} \label{sec:train-random-gap}

The train--random gap measures NW training accuracy beyond the random-suffix fitting reference,
\[
    G(t) = A_T(t) - A_R(t).
\]
The \emph{early transition} is the initial rise of NW validation accuracy from zero. Figure~\ref{fig:figure3}(d--f) shows the gap becoming positive alongside validation, as matrix accuracy exceeds suffix accuracy.

During the early transition, the empirical relation $A_T(t)\approx A_V(t)+A_R(t)$ is consistent with contributions from both transferable rule learning and example-specific fitting.
The agreement between $A_T(t)-A_R(t)$ and $A_V(t)$ suggests that the rule improving validation predictions also helps fit the training set. Generalization therefore emerges while example-specific fitting is still incomplete, rather than beginning only after memorization is finished. Because $A_R$ measures a separate target, the relation provides a diagnostic of the two contributions rather than an exact partition of training examples. Once validation accuracy is high, $A_T$ saturates while $A_R$ continues to rise (Figure~\ref{fig:figure3}(b--c)), so the additive account describes the early transition, not the entire learning trajectory.

These observations suggest that acquiring shared predictions and fitting example-specific associations can contribute simultaneously to training progress. The increasing cost of suffix memorization motivates investigating whether analogous residual fitting costs contribute to the original NW slowdown. The probe alone does not determine why the response differs between NW and multiplication. Section~\ref{sec:eca} therefore examines \emph{task learnability}, asking what makes a rule easier to acquire and how larger datasets can weaken that advantage.

\section{Why More Data Slows Learning: Task Learnability} \label{sec:eca}

We investigate the large-data slowdown through task learnability. NW and multiplication differ in both their rules and output formats. Elementary cellular automata (ECA) repeatedly update each binary cell using its own state and those of its two neighbors. ECA let us compare rules at fixed output length, recurrence structure, and training settings. We test whether lower-order signals make a rule easier to learn and whether larger datasets weaken those signals.

\subsection{Learning Regimes in a Controlled ECA Setting}
\label{sec:eca-regimes}

In an elementary cellular automaton (ECA), each cell updates its binary state based only on its own current state and those of the two adjacent cells. Let $x_i^t\in\{0,1\}$ denote the state of cell $i$ at time $t$, and let $x^t=(x_1^t,\ldots,x_L^t)$ denote the full row of states. Starting from an initial row $x^0$, the same rule $f_r:\{0,1\}^3\to\{0,1\}$~\citep{wolfram1983statistical} updates all cells simultaneously,
\begin{equation*}
    x_i^t=f_r(x_{i-1}^{t-1},x_i^{t-1},x_{i+1}^{t-1}).
\end{equation*}
We call the three preceding-row states $(x_{i-1}^{t-1},x_i^{t-1},x_{i+1}^{t-1})$ the \emph{parents} of $x_i^t$. Cyclic boundaries make the first and last cells adjacent. The model receives an initial row $x^0$ of $L=20$ cells and generates the next $T=8$ rows $x^1,\ldots,x^8$, concatenated with row separators. Under the uniform population distribution, the initial states $x_i^0$ are independent Bernoulli$(1/2)$ variables. We use the same initial rows for all three rules. Appendix~\ref{app:eca-details} describes the model, optimizer, dataset, and evaluation.

\begin{figure}[tp]
    \centering
    \includegraphics[width=\linewidth]{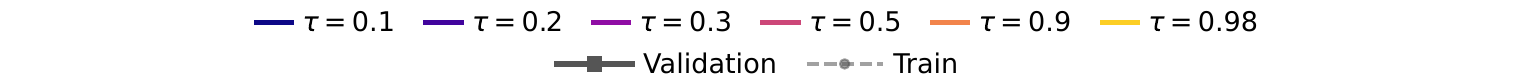}\par\vspace{2pt}
    \begin{subfigure}[t]{0.34\linewidth}
        \centering
        \includegraphics[width=\linewidth]{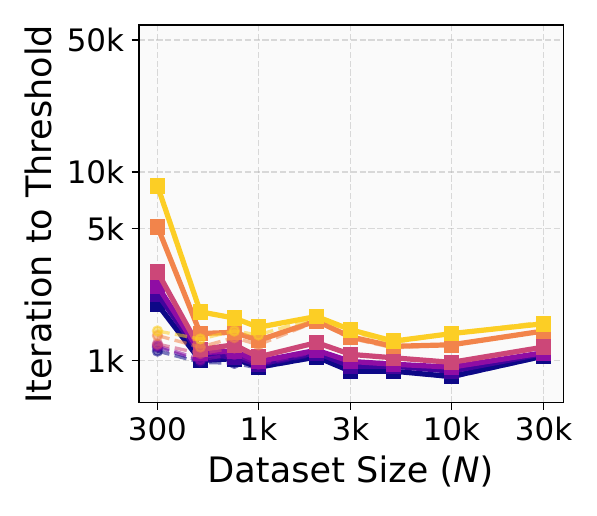}
        \caption{Rule 110.}
        \label{fig:figure4a}
    \end{subfigure}%
    \hspace{6pt}%
    \begin{subfigure}[t]{0.28\linewidth}
        \centering
        \includegraphics[width=\linewidth]{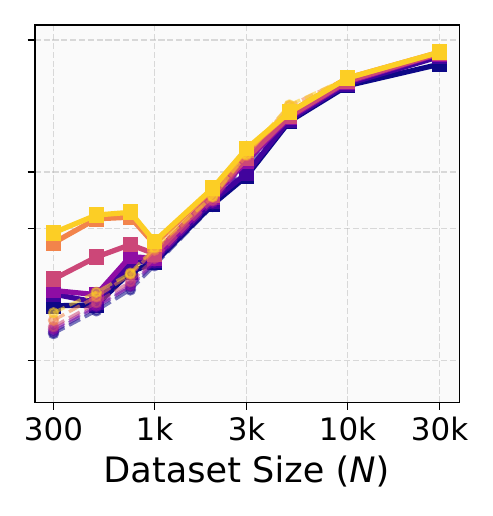}
        \caption{Rule 90.}
        \label{fig:figure4b}
    \end{subfigure}%
    \hspace{6pt}%
    \begin{subfigure}[t]{0.28\linewidth}
        \centering
        \includegraphics[width=\linewidth]{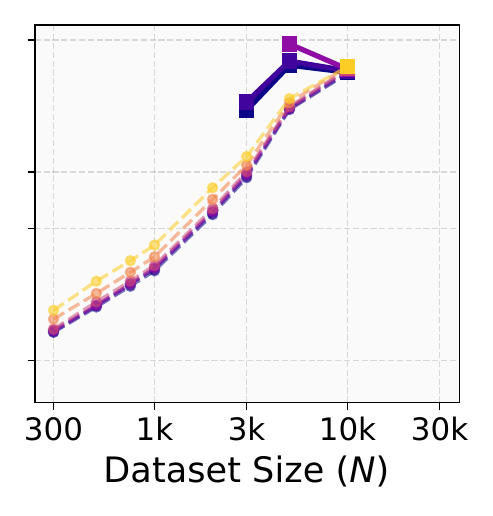}
        \caption{Rule 150.}
        \label{fig:figure4c}
    \end{subfigure}
    \caption{Dataset-size regimes in ECA. Rules~110, 90, and 150 use the same input and output format and the same training settings. Colors denote exact-trajectory thresholds. Solid-square curves show validation crossings, and dashed-circle curves show training-dataset diagnostic crossings. Rule~110 quickly reaches a near-flat update requirement, Rule~90 exhibits a large-data slowdown after the training--validation delay narrows, and Rule~150 remains difficult to generalize within the budget.}
    \label{fig:figure4}
\end{figure}

Rule~110 is asymmetric and nonlinear, whereas Rules~150 and 90 compute parity over three and two parents, respectively. Rule~90 uses the left and right parents. We describe their observed learning regimes as easy, hard, and intermediate, respectively. These rules exhibit distinct learning regimes despite identical output lengths, recurrence structures, and training settings.

\paragraph{Easy: initial speedup, then a plateau.} Rule~110 generalizes in fewer updates as the dataset first grows (Figure~\ref{fig:figure4}(a)). At larger dataset sizes, the required updates remain roughly constant, and training and validation improve on similar timescales. The pattern is consistent with grokking studies in which additional data shortens the delay between fitting and generalization~\citep{power2022grokking,liu2022towards,zhu2024critical}.

\paragraph{Hard: little generalization within the budget.} Rule~150 (Figure~\ref{fig:figure4}(c)) fits the training diagnostic at smaller dataset sizes but rarely generalizes. With more data, it takes longer to reach the training thresholds and eventually fails to reach them within the budget. Thus, a short and exactly specified parity rule can still be hard for the models to learn.

\paragraph{Intermediate: narrowing train--validation delay, then slowdown.} Rule~90 (Figure~\ref{fig:figure4}(b)) exhibits two contrasting responses within the same learnable task. At smaller dataset sizes, training accuracy rises well before validation, exhibiting a delay characteristic of grokking~\citep{power2022grokking,liu2022towards}. At larger dataset sizes, the mean recorded training and validation crossing times nearly coincide and increase together with $N$. Unlike Rule~110, Rule~90 shows no plateau in acquisition time as $N$ grows. It also generalizes across a wider range of dataset sizes than Rule~150. At fixed output length and recurrence structure, we next examine which rule properties make learning easier.

\subsection{Lower-Order Signals Make Rules Easier to Learn} \label{sec:eca-signals}

Useful prediction can begin before the model learns the full rule. The model can first capture patterns that require little input information, such as an imbalance in target frequencies or a predictive relationship involving only a few inputs. Such patterns need not determine the output to improve accuracy and reduce loss relative to uniform guessing. A \emph{signal} is the predictive structure that makes such improvement possible. We quantify the \emph{$k$-order signal}, $G_k(P)$, as the largest reduction in target uncertainty an ideal predictor can obtain under $P$ using at most $k$ parent bits.

At a fixed position $i$ and transition $t$, let $Y=x_i^t$ denote the target and $\mathcal Y$ its set of possible labels. Let $P$ be the distribution of the preceding row. For $U\subseteq\{i-1,i,i+1\}$, write $X_U=(x_j^{t-1})_{j\in U}$ for the observed parents, and let $\mathcal U_k$ contain all such subsets with at most $k$ elements, including the empty set. The $k$-order signal is
\begin{equation*}
    G_k(P)=\log_2|\mathcal Y|-\min_{U\in\mathcal U_k}H_P(Y\mid X_U)
          =G_0(P)+\max_{U\in\mathcal U_k}I_P(Y;X_U).
\end{equation*}
The reference predictor assigns equal probability to every possible label, giving an uncertainty of $\log_2|\mathcal Y|$ bits. In our ECA task, $\mathcal Y=\{0,1\}$, so this reference uncertainty is $\log_2 2=1$ bit. The conditional entropy $H_P(Y\mid X_U)$ measures the uncertainty remaining after observing the parents in $U$, and the minimum selects the most informative allowed subset. Thus, $G_k(P)$ measures how much target uncertainty can be reduced using at most $k$ parents. Equivalently, it combines the output-bias signal $G_0(P)=\log_2|\mathcal Y|-H_P(Y)$, arising from unequal target frequencies, with the additional signal from the observed parents, measured by mutual information $I_P(Y;X_U)$.

The order $k$ bounds the number of observed parents, rather than isolating a pure $k$-way interaction. The quantities describe available signals, not signals necessarily learned by the model. Every ECA rule has $G_3(P)=1$ because its three parents determine the output exactly. The full three-parent signal therefore does not distinguish the rules. Signals available from fewer or no parent bits are \emph{lower-order signals}, which can support useful prediction before the full rule is known. Comparing $G_0$, $G_1$, and $G_2$ reveals whether the rules offer similarly easy ways to begin learning.

\begingroup
\setlength{\intextsep}{4pt}
\setlength{\columnsep}{10pt}
\begin{wrapfigure}{r}{0.43\linewidth}
\centering
    \includegraphics[width=\linewidth]{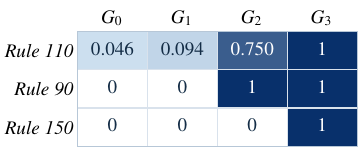}
    \caption{Uniform one-step $k$-order signals. Darker blue indicates larger values, with zero shown in white.}
    \label{fig:figure5}
\end{wrapfigure}

Rule~110 offers useful predictions from fewer parents than Rules~90 and 150 (Figure~\ref{fig:figure5}). Under uniform one-step inputs, Rule~110 has positive $G_0$ and $G_1$, although no pair fully determines the output ($G_2<1$). Rule~90 has the larger second-order signal, $G_2=1$, but $G_0=G_1=0$, and Rule~150 requires all three parents for useful prediction. Perfect prediction from two parents does not make Rule~90 easier than Rule~110. Lower-order signals can matter more for acquisition than a stronger signal that requires combining parents. For staircase targets, gradient-based learning can acquire higher-order features by building on lower-order features learned earlier~\citep{abbe2021staircase,abbe2023leap}. The staircase perspective suggests that Rule~110 is easier because useful lower-order predictions can aid learning before relationships involving more parents are acquired. Uniform Rules~90 and 150 require multiple parents to improve on chance.
\par
\endgroup

\setlength{\textfloatsep}{7pt plus 2pt minus 2pt}
\subsection{Adding Lower-Order Signals Accelerates Rule 90 Acquisition}
\label{sec:eca-biased}

\begin{figure}[!t]
    \centering
    \begin{subfigure}[t]{0.455\linewidth}
        \centering
        \includegraphics[width=\linewidth]{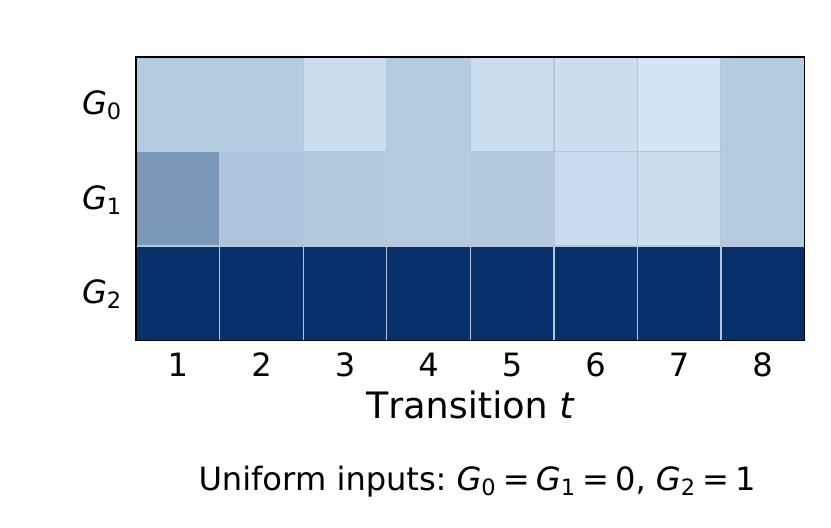}
        \caption{Inputs with 13--15 ones.}
        \label{fig:figure6a}
    \end{subfigure}
    \hfill
    \begin{subfigure}[t]{0.455\linewidth}
        \centering
        \includegraphics[width=\linewidth]{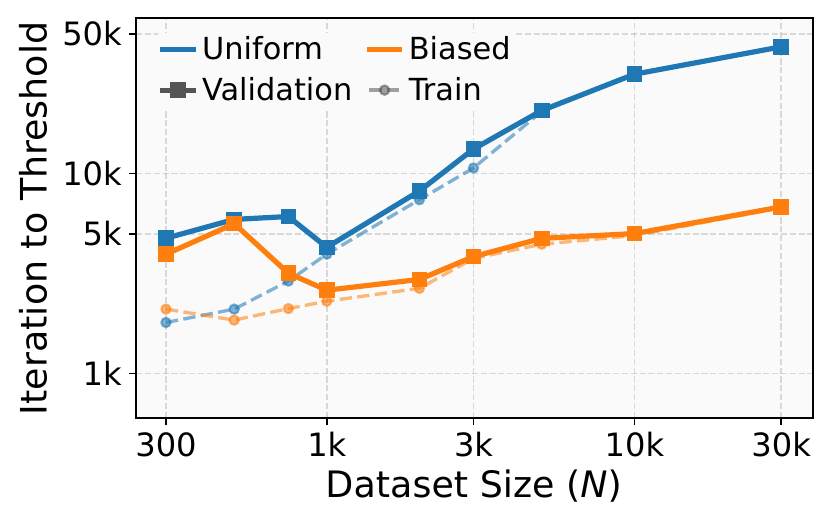}
        \caption{Rule 90, $\tau=0.98$.}
        \label{fig:figure6b}
    \end{subfigure}
    \caption{Biased training reduces Rule~90's slowdown. \textbf{(a)}: $k$-order signals across eight transitions for inputs with 13--15 ones. Biased inputs introduce lower-order signals absent under uniform inputs ($G_0=G_1=0$). White denotes zero and darker blue denotes larger $G_k$. \textbf{(b)}: mean updates to reach $0.98$ exact eight-row accuracy. Validation remains unchanged and uniform. Solid squares and dashed circles mark validation and training-dataset diagnostic crossings, respectively. Within each condition, both means use runs that reach validation accuracy $0.98$.}
    \label{fig:figure6}
\end{figure}

We test whether adding lower-order signals makes Rule~90 faster to learn. We hold the XOR rule, model, and optimization settings fixed and manipulate the distribution of initial rows. Uniformly sampled rows serve as the baseline, while biased sampling restricts the 20-bit initial rows to states containing 13--15 ones. The biased distribution introduces positive $G_0$ and $G_1$ without changing the rule to be learned (Figure~\ref{fig:figure6}(a)). Both training conditions use the same uniform validation set, so the comparison tests whether providing lower-order signals during training accelerates acquisition on the original task. Appendix~\ref{app:eca-details} details the construction and signal calculation.

Biased training reaches the validation threshold in substantially fewer updates at larger dataset sizes (Figure~\ref{fig:figure6}(b)).
Since the biased distribution introduces lower-order signals, this result supports a role for such signals in rule acquisition on uniform validation inputs. The bias here was deliberate. We next examine signals introduced by random sampling, which can change with dataset size.

\subsection{\shin{More Data, Weaker Signals, Slower Learning}} \label{sec:eca-empirical}

\begin{figure}[!t]
    \centering
    \begin{subfigure}[t]{0.455\linewidth}
        \centering
        \includegraphics[width=\linewidth]{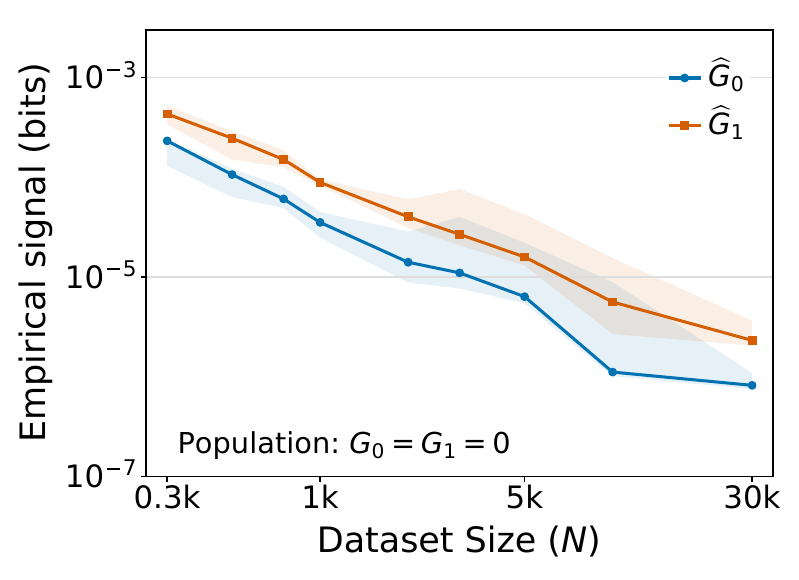}
        \caption{Empirical lower-order signals.}
        \label{fig:figure7a}
    \end{subfigure}
    \hfill
    \begin{subfigure}[t]{0.455\linewidth}
        \centering
        \includegraphics[width=\linewidth]{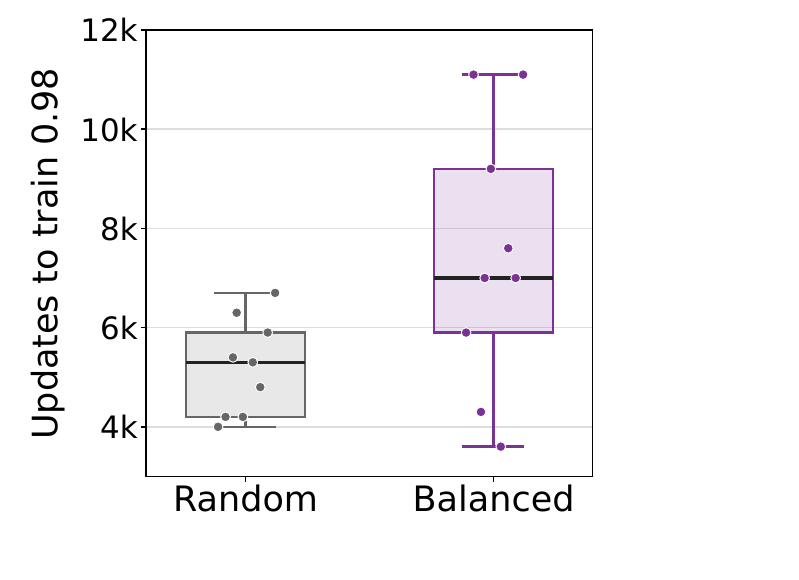}
        \caption{Matched $N=2$k datasets.}
        \label{fig:figure7b}
    \end{subfigure}
    \caption{Empirical signals and training-set fitting in Rule~90.
    \textbf{(a)}: empirical $G_0$ and $G_1$ approach their population value of zero as dataset size grows.
    Lines and bands show the median and range over three datasets after averaging over the eight transitions.
    \textbf{(b)}: number of updates to reach $0.98$ training accuracy at $N=2{,}000$.
    Targets and first-transition $G_0=0$ are matched, but $G_1>0$ for Random and $G_1=0$ for Balanced.
    Balanced fits more slowly on average. Boxes show medians and interquartile ranges over nine runs. Dots show individual runs.}
    \label{fig:figure7}
\end{figure}

Smaller datasets can supply lower-order signals absent from the population. Under uniform initial cells, Rule~90 has $G_0(P)=G_1(P)=0$ for the transitions studied. A sampled dataset's empirical distribution $\widehat P_N$ need not match the population exactly. Chance deviations create output imbalance and single-parent dependence, producing positive $\widehat G_k=G_k(\widehat P_N)$ for $k=0,1$~\citep{paninski2003estimation}. Such deviations are typically larger in smaller samples. As $N$ grows, the empirical distribution approaches the population and the signals decrease toward zero (Figure~\ref{fig:figure7}(a)). The accidental patterns can help early training predictions even though they need not generalize to the population.

We test whether the empirical signals affect learning by comparing \emph{Random} and \emph{Balanced} datasets at $N=2{,}000$. The target trajectories and their frequencies are identical, and only the initial rows differ. Both datasets have first-transition $G_0=0$, but Random retains $G_1>0$ while Balanced has $G_1=0$. Balanced takes more updates, on average, to reach training accuracy $0.98$ (Figure~\ref{fig:figure7}(b)). The comparison supports the learning benefit of empirical lower-order signals, although balancing also changes other input statistics. Appendix~\ref{app:eca-details} gives the construction.

More data can therefore describe the population better while weakening empirical signals that make learning easier.
In Rule~90, weakening lower-order signals helps explain why larger datasets take longer to learn.
The NW suffix probe provides complementary evidence that fitting arbitrary example-specific associations becomes more costly as the dataset grows (Section~\ref{sec:mechanism}). Together, these experiments suggest that weakening empirical signals and increasing fitting costs may contribute to the observed slowdowns.

\section{Discussion and Limitations} \label{sec:discussion}

\paragraph{What the evidence supports.}
The large-data slowdown is not limited to full NW tables or a single training configuration. Depth, batch-size, weight-decay, and cell-accuracy checks support the slowdown across multiple settings (Appendices~\ref{app:depth} and~\ref{app:additional-robustness}). Edit-distance tables and partial NW outputs retain a sweet spot, while scratchpad multiplication also slows at larger dataset sizes (Appendix~\ref{app:additional-tasks}). In ECA, rules with identical output length and recurrence structure exhibit different learning regimes. Output format alone therefore cannot explain the contrast. Lower-order signals help explain differences in task learnability and why larger Rule~90 datasets can learn more slowly.

\paragraph{When can smaller datasets help?}
Our results suggest that the benefits of additional examples can be offset by changes in the predictive signals available during training and by the cost of fitting example-specific associations. 
In Rule~90, sampling fluctuations supply lower-order signals that weaken as the dataset grows. The NW suffix probe shows that memorizing example-specific associations takes longer with more examples. The account connects to the benefits of repeated examples reported by \citet{charton2024repeated} and the sampling-bias explanation of \citet{liu2026lessdata}. Useful biases need not arise only by chance. Deliberately biased Rule~90 training accelerates learning on an unchanged uniform validation distribution (Section~\ref{sec:eca-biased}). The result motivates testing signal-rich synthetic datasets and curricula that gradually broaden the training distribution in code and reasoning tasks. Dataset size and composition should therefore be treated as training choices, rather than assuming that more distinct examples always make learning faster.

\paragraph{Limitations and open questions.}
We do not yet predict which task properties place a problem in the intermediate regime or determine its optimal dataset size.
The lower-order-signal account is tested in Rule~90, but comparable signals have not been measured directly in NW.
Defining those signals and identifying how rule learning and example-specific fitting interact inside the model would test the proposed explanation more directly.
Experiments with larger models and variable-length inputs are also needed to establish how broadly the findings apply.

\section{Conclusion}
\label{sec:conclusion}
NW reaches high validation accuracy fastest at an intermediate dataset size, even though adding examples can initially accelerate training-set fitting. Controlled ECA experiments connect the differing data-size responses to task learnability. Biased Rule~90 training that introduces lower-order signals accelerates acquisition on uniform validation inputs, while empirical signals from random sampling weaken with dataset size. Together, the Rule~90 interventions and the NW random-suffix diagnostic suggest complementary pressures that may contribute to the large-data slowdown. More examples can help reveal a rule without reducing the number of updates needed to learn it.

\clearpage
\begin{ack}
This work was supported by the National Research Foundation of Korea (NRF) grant funded by the Korean government (MSIT) in 2026 (Grant No.~RS-2026-25612350, Project Title: ``Development of an End-to-End Design and Validation Framework for Mathematically Principled Next-Generation AI Architectures and Learning Algorithms''), the National Research Foundation of Korea (NRF) grant funded by the Korean government (MSIT) (No.~RS-2024-00410005), and the Pioneer Research Center Program through the National Research Foundation of Korea funded by the Ministry of Science, ICT \& Future Planning (RS-2022-NR067569).

\paragraph{Competing interests.}
The authors declare that, during the 36 months preceding submission, they had no financial relationships outside the submitted work that could reasonably be \mbox{perceived as influencing this research.}
\end{ack}

\bibliographystyle{plainnat}
\bibliography{references}

\clearpage
\appendix

\setcounter{topnumber}{2}
\setcounter{bottomnumber}{1}
\setcounter{totalnumber}{3}

\section{Task Details}
\label{app:task_details}

\subsection{Needleman--Wunsch Task Details}
\label{app:nw_details}

We use fixed-length Needleman--Wunsch matrix prediction as a controlled structured-output task. Each input consists of two sequences $x=(x_1,\ldots,x_m)$ and $y=(y_1,\ldots,y_m)$ over a finite alphabet $\mathcal{A}$. The target is the full dynamic-programming score matrix $M \in \mathbb{Z}^{(m+1)\times(m+1)}$ computed by the Needleman--Wunsch recurrence. The main experiments use $m=5$ and $|\mathcal{A}|=4$.

The score matrix is initialized by
\[
M_{0,0}=0,\qquad
M_{i,0}=i\cdot g,\qquad
M_{0,j}=j\cdot g,
\]
where $g$ is the gap penalty. For $i,j\geq 1$, the recurrence is
\[
M_{i,j}
=
\max\left\{
M_{i-1,j-1}+s(x_i,y_j),
M_{i-1,j}+g,
M_{i,j-1}+g
\right\},
\]
where $s(x_i,y_j)$ is the match/mismatch score. We use $s(a,b)=5 \; (\text{match}), -4\; (\text{mismatch})$ and gap penalty $g=-5$ throughout the main experiments.

The model predicts the score matrix as a sequence of discrete output tokens. Unless otherwise stated, we evaluate exact full-matrix accuracy: an example is counted as correct only when all predicted matrix entries match the Needleman--Wunsch target matrix. Cell-level accuracy is used only for diagnostic analyses and is not the main convergence metric.

For each dataset size $N$, we sample $N$ training examples from the finite task universe and evaluate on a fixed held-out validation set. Unless otherwise stated, the validation set is held fixed across runs for the same task configuration. Dataset sizes are swept over $N\in \{200, 500, 1k, 2k, 3k, 4k, 5k, 10k, 50k, 100k\}$. 

\subsection{Addition and Multiplication Baselines}
\label{app:arith_details}

We use decimal addition and multiplication as short-output algorithmic baselines. These tasks are
included to test whether exact-match thresholding alone produces an intermediate-data sweet spot.
The main claims of the paper concern Needleman--Wunsch matrix generation; the arithmetic tasks
serve as contrast controls.

For \(d\)-digit addition, each input is a pair of nonnegative decimal integers
\((a,b)\), where \(a,b\in\{0,\ldots,10^d-1\}\), and the target is the exact decimal representation of
\(a+b\). For \(d\)-digit multiplication, the input is again a pair \((a,b)\), with
\(a,b\in\{0,\ldots,10^d-1\}\), and the target is the exact decimal representation of \(a\cdot b\).
The operand lengths are \(d=3\) for multiplication (Figure~\ref{fig:figure1a}) and \(d=10\) for \mbox{the addition task (Figure~\ref{fig:figure15}).}

Inputs are blank-padded to length \(d\). Targets are first written in canonical decimal form without leading zeros and then blank-padded to the maximum target length for batching and autoregressive prediction. Exact-match accuracy is computed on the full padded target sequence.

For arithmetic baselines, training examples are sampled from the finite input universe and validation
examples are held out from the same task family. The multiplication sweep uses
\[
N \in \{500, 1\mathrm{k},5\mathrm{k},10\mathrm{k},20\mathrm{k},30\mathrm{k},40\mathrm{k},50\mathrm{k},100\mathrm{k},200\mathrm{k},300\mathrm{k},400\mathrm{k},500\mathrm{k}\}.
\]
The addition sweep uses
\[
N \in \{500, 600,700,800,900,1\mathrm{k},5\mathrm{k},10\mathrm{k},20\mathrm{k},30\mathrm{k},40\mathrm{k},50\mathrm{k},100\mathrm{k},200\mathrm{k},300\mathrm{k},400\mathrm{k},500\mathrm{k}\}.
\]
Optimizer, model-size, threshold-crossing, and censoring conventions follow NW unless noted.

\section{Full Model and Training Details}
\label{app:training_details}

\subsection{Dataset sweep and model}

For NW at length $L$, the finite task universe has size $|\mathcal{D}_{\mathrm{full}}|=|\Sigma|^{2L}=4^{2L}$. At $L=5$, this gives $4^{10}=1{,}048{,}576$ possible input pairs. For each dataset size $N$, we sample a training set $\mathcal{D}_N\subset\mathcal{D}_{\mathrm{full}}$ and evaluate on held-out examples from the same universe. The NW sweep uses
\[
N\in\{200,500,1\mathrm{k},2\mathrm{k},3\mathrm{k},4\mathrm{k},5\mathrm{k},10\mathrm{k},50\mathrm{k},100\mathrm{k}\}.
\]

We train decoder-only Transformers \citep{vaswani2017attention}. A depth-$D$ model has $D$ layers, $D$ attention heads, embedding dimension $64D$, and feed-forward width $256D$. The main depth sweep uses $D\in\{3,4,5,6\}$. Models are trained with AdamW \citep{loshchilov2019decoupled} and a cosine learning-rate schedule from $10^{-3}$ to $10^{-4}$, with $\beta_1=0.9$, $\beta_2=0.99$, and weight decay $0.1$. The effective batch size is $160$, obtained by gradient accumulation, and each run is trained for at most $50{,}000$ optimizer updates with evaluation every $100$ updates. Batch size, optimizer settings, tokenization, and evaluation frequency are held fixed within each task sweep. We treat all non-crossing runs as censored rather than assigning them finite crossing times. Each task and dataset-size configuration is run with five random seeds unless otherwise stated.

The baseline effective batch of 160 combines eight sequences per microbatch with 20 gradient-accumulation steps. Models are trained from random initialization. Shaded regions report one standard deviation across runs with a recorded crossing at the plotted threshold.

If a run does not reach the threshold within $50{,}000$ updates, we treat it as censored and do not assign it a finite crossing time. Figure~\ref{fig:figure1} uses the representative $D=6$ setting, Figure~\ref{fig:figure2} shows $D=4,5$, and Appendix~\ref{app:depth} provides the full depth sweep.

\paragraph{Compute resources.}
The original experiments were run on NVIDIA RTX 3090 GPUs with 24 GB memory. A single NW run at L=5 and depth D=6 takes approximately 10 GPU-hours for 50,000 optimizer updates. The original experimental suite used approximately 2,600 GPU-hours, including the main dataset-size sweep, depth sweep, sequence-length robustness experiments, and random-suffix probes. Preliminary or failed runs used approximately 1,077 additional GPU-hours.

\section{Robustness and Measurement Scope}
\label{sec:robustness}
\label{app:measurement-scope}

The robustness evidence includes the original depth and sequence-length sweeps, seed variability, epoch normalization, and full random-suffix trajectories (Appendices~\ref{app:depth}--\ref{app:suffix}). Further checks vary effective batch size, weight decay, and model depth, and compare cell accuracy with exact match (Appendix~\ref{app:additional-robustness}). Additional algorithmic tasks are reported in Appendix~\ref{app:additional-tasks}.

\paragraph{Depth robustness.}
Figure~\ref{fig:figure2} repeats the NW sweep across depths. The qualitative pattern is stable: validation crossing time is non-monotonic in $N$, and train crossing time can decrease near the onset of weak validation competence. 
The optimal dataset size shifts with depth, but the two findings do not rely on a single representative model. 

\paragraph{Threshold robustness.}
The use of multiple thresholds is essential. A single strict threshold would show only the late stage of learning. The lower thresholds reveal that validation competence emerges gradually and can precede strict train convergence. 
These crossings are consistent with transferable structure being learned while training-set fitting remains incomplete. The strict threshold $\tau=0.98$ is used for the headline sweet spot because it measures high exact-match accuracy; weaker thresholds diagnose the transition into that regime.

For the NW and arithmetic plots with recorded cutoffs spaced by $1/80$, the displayed crossing is selected at the smallest recorded cutoff at least as high as $\tau$. Curves labeled $\tau=0.98$ therefore use the recorded $0.9875$ crossing. These cutoffs are equivalent for exact-match evaluation on 80 examples, but need not be equivalent at other evaluation sizes or for cell accuracy.

NW crossing times are the first evaluated checkpoints that meet the threshold, with evaluation every 100 optimizer updates. A recorded crossing does not require accuracy to remain above the threshold at later checkpoints.

\paragraph{Update-based scope.}
The slowdown we report is specifically update-based. One optimizer update sees a fixed batch size, not a fixed fraction of the dataset. As $N$ grows, the same number of updates covers fewer epochs. Thus, our result should not be read as saying that larger datasets are intrinsically worse, or that they reduce asymptotic accuracy. It says that, for strict exact-match NW generation under the training protocol studied here, the fastest path in gradient-update count occurs at an intermediate dataset size. Epoch-normalized curves answer a different question: how many passes over the data are needed, rather than how many parameter updates are needed.

\paragraph{Censoring and interpretation.}
Censoring is applied to individual runs: a run that does not reach a threshold within the budget contributes no finite crossing time. For the NW and arithmetic sweeps, training and validation means at each threshold use the runs with recorded crossings on the corresponding split. A dataset size remains eligible if at least one run reaches the threshold. Thus, the reported means condition on within-budget threshold attainment and do not include the probability of attainment. The critical size $N_c(\tau)$ is the smallest tested size with a recorded validation crossing, and the optimal size $N^\star(\tau)$ minimizes the mean among sizes with such crossings. The observed sweet spot compares intermediate and larger datasets with recorded crossings; its interpretation is conditional on the contributing runs.

\FloatBarrier
\section{Robustness Across Depth} \label{app:depth}

\begin{figure}[!htbp]
    \centering
    \includegraphics[width=\linewidth]{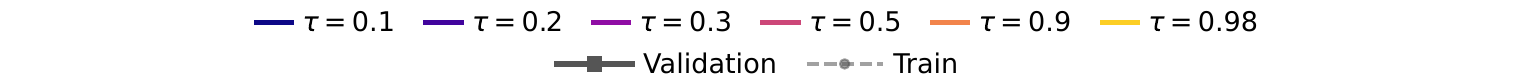}\par\vspace{2pt}
    \begin{subfigure}[t]{0.49\linewidth}
        \centering
        \includegraphics[width=\linewidth]{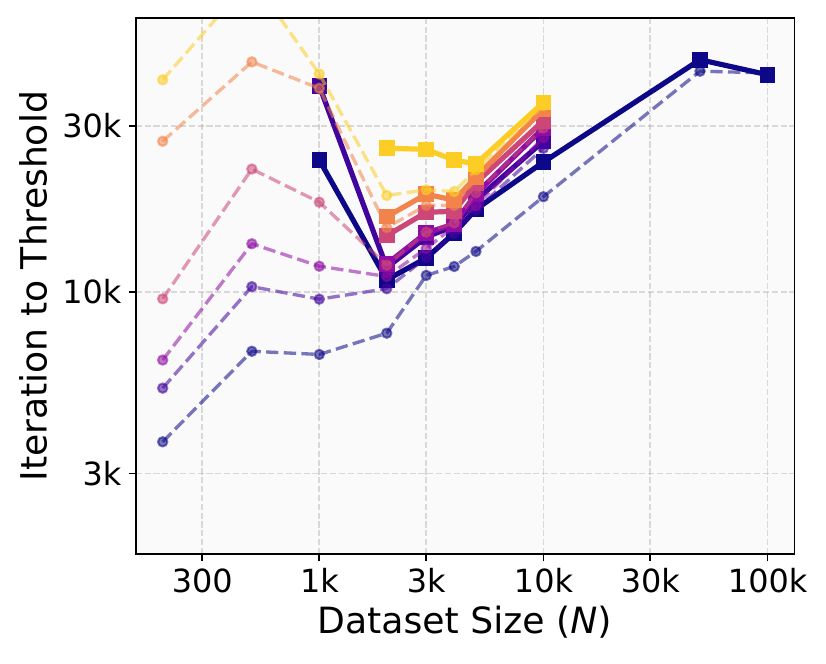}
        \caption{Depth 3.}
        \label{fig:figure8a}
    \end{subfigure}
    \hfill
    \begin{subfigure}[t]{0.49\linewidth}
        \centering
        \includegraphics[width=\linewidth]{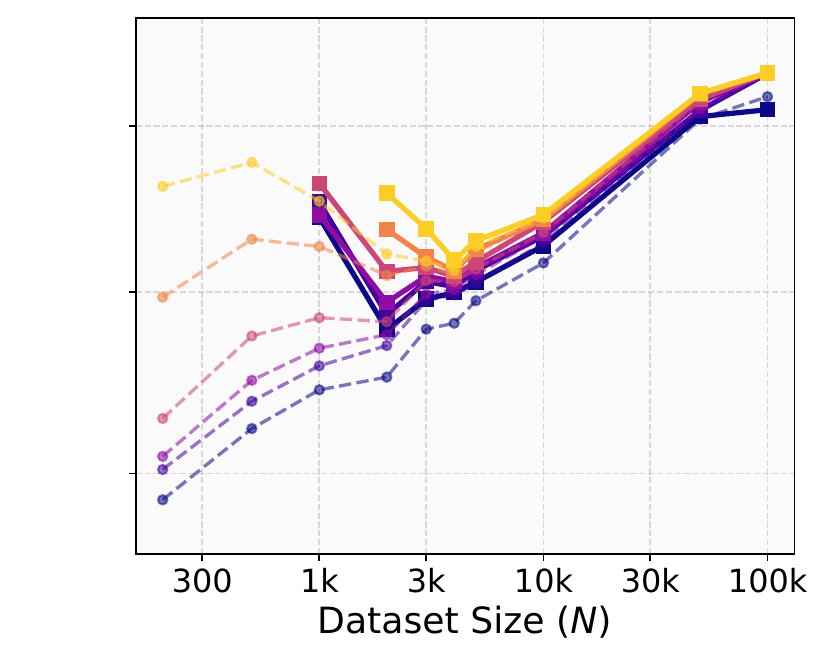}
        \caption{Depth 4.}
        \label{fig:figure8b}
    \end{subfigure}
    \par\vspace{4pt}
    \begin{subfigure}[t]{0.49\linewidth}
        \centering
        \includegraphics[width=\linewidth]{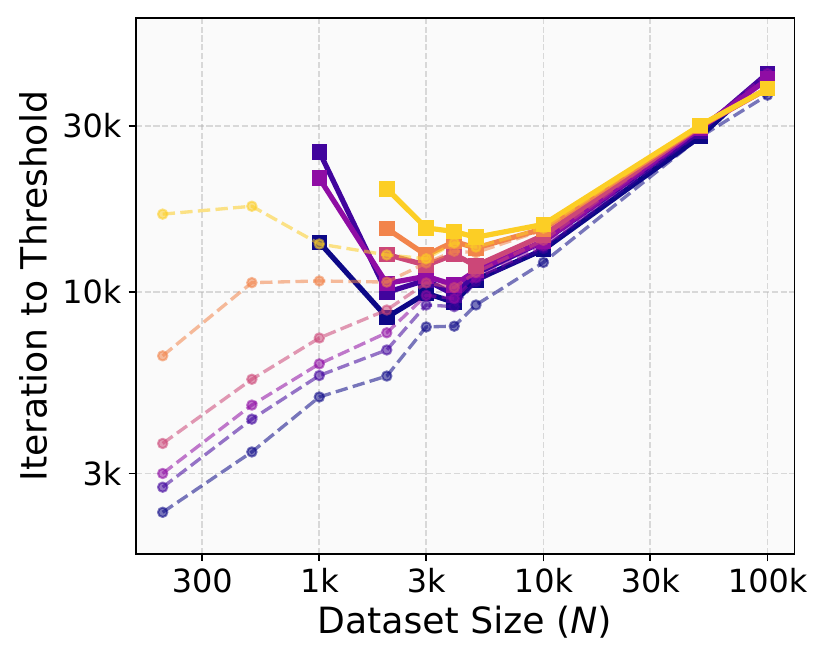}
        \caption{Depth 5.}
        \label{fig:figure8c}
    \end{subfigure}
    \hfill
    \begin{subfigure}[t]{0.49\linewidth}
        \centering
        \includegraphics[width=\linewidth]{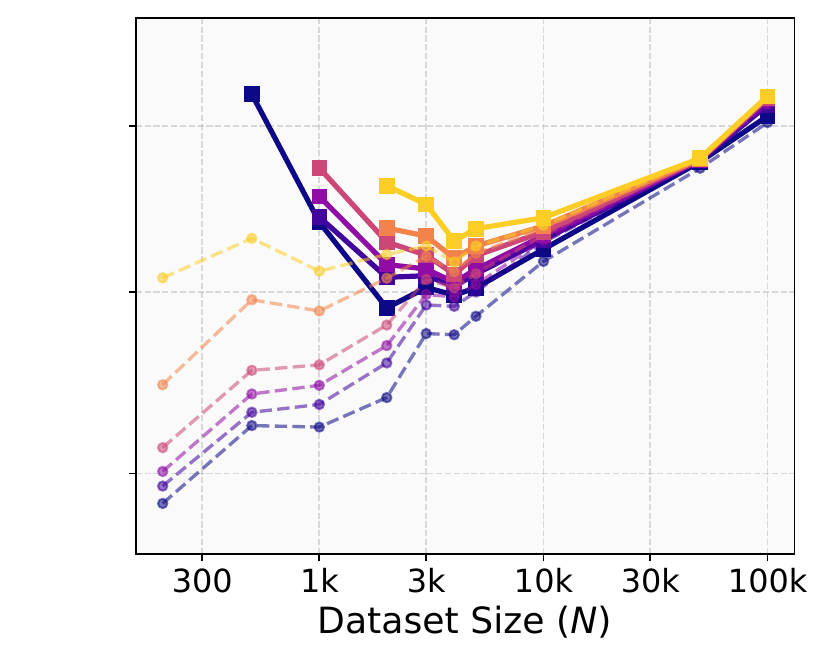}
        \caption{Depth 6.}
        \label{fig:figure8d}
    \end{subfigure}
    \caption{
    NW multi-threshold sweeps across depths \(D\in\{3,4,5,6\}\) at \(L=5\). Colors denote exact-match thresholds, solid lines show validation crossings, and dashed lines show training crossings. Across depths, validation crossing time is non-monotonic in dataset size, giving an intermediate-data sweet spot. In the same dataset-size band where weak validation thresholds first become reachable, training crossings can also become faster as \(N\) increases, supporting the partial-rule interpretation.
    }
    \label{fig:figure8}
\end{figure}

We test whether Findings 1 and 2 are specific to the representative $D=6$, $L=5$ setting or whether they reproduce across model depth. Figure~\ref{fig:figure8} reports the multi-threshold trajectories of Figure~\ref{fig:figure1b} for $D \in \{3,4,5,6\}$ at $L=5$.
The qualitative pattern is preserved across depths: validation crossing time is non-monotonic in $N$ (Finding 2), and the strict training crossing $I_T(0.98;N)$ can decrease near the onset of weak validation competence (Finding 1). The optimal dataset size varies with model depth.

\begin{figure}[!htbp]
    \centering
    \includegraphics[width=\linewidth]{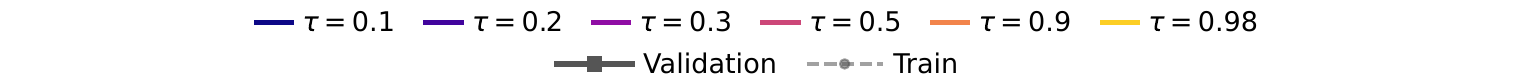}\par\vspace{2pt}
    \begin{subfigure}[t]{0.49\linewidth}
        \centering
        \includegraphics[width=\linewidth]{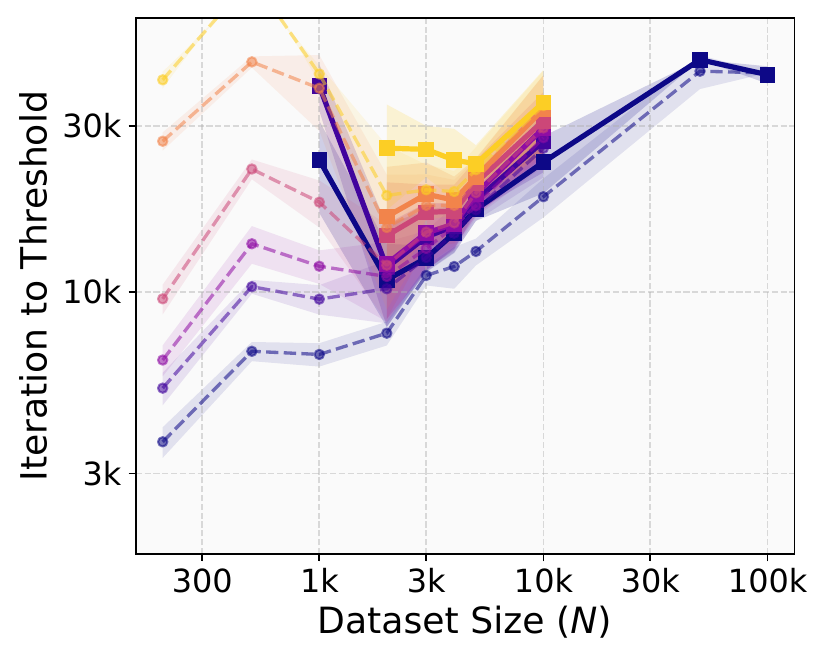}
        \caption{Depth 3.}
        \label{fig:figure9a}
    \end{subfigure}
    \hfill
    \begin{subfigure}[t]{0.49\linewidth}
        \centering
        \includegraphics[width=\linewidth]{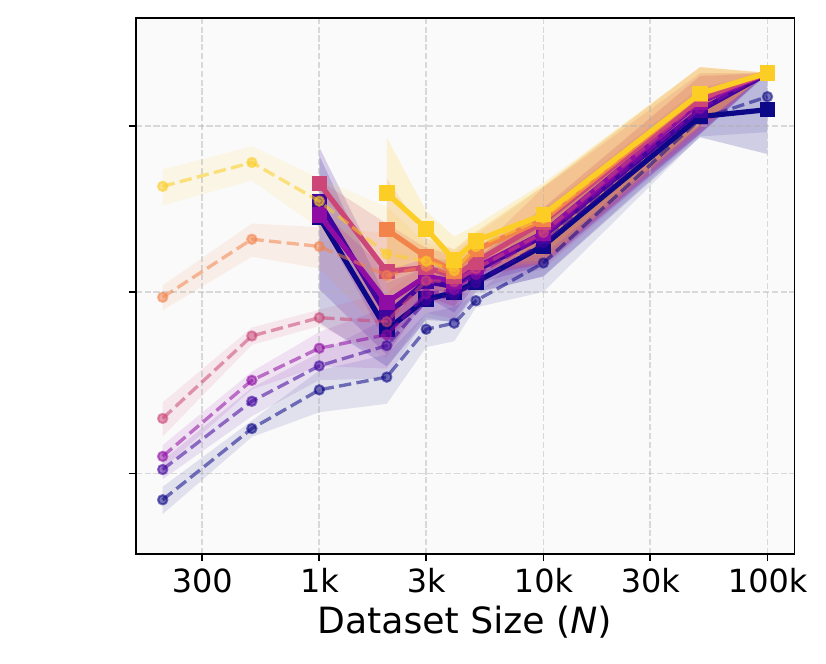}
        \caption{Depth 4.}
        \label{fig:figure9b}
    \end{subfigure}
    \par\vspace{4pt}
    \begin{subfigure}[t]{0.49\linewidth}
        \centering
        \includegraphics[width=\linewidth]{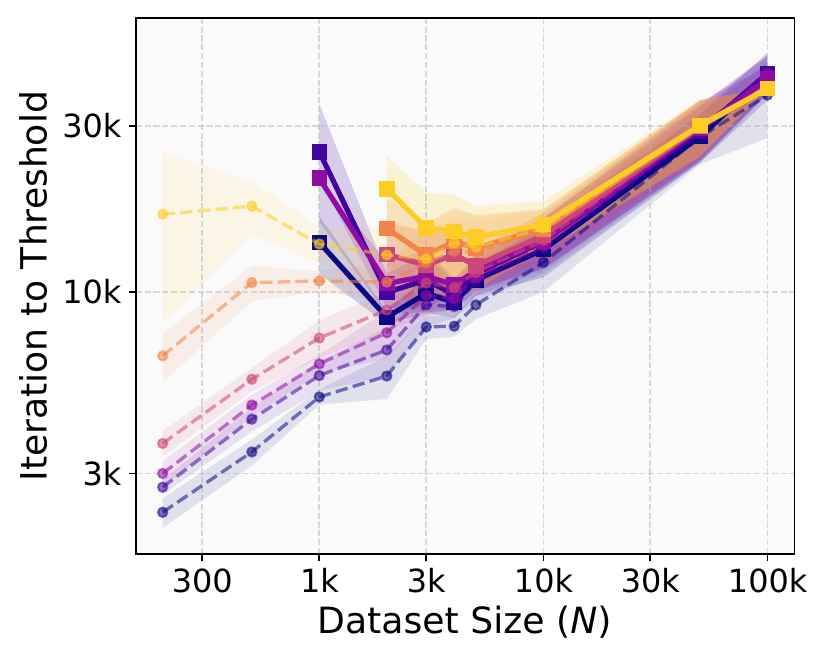}
        \caption{Depth 5.}
        \label{fig:figure9c}
    \end{subfigure}
    \hfill
    \begin{subfigure}[t]{0.49\linewidth}
        \centering
        \includegraphics[width=\linewidth]{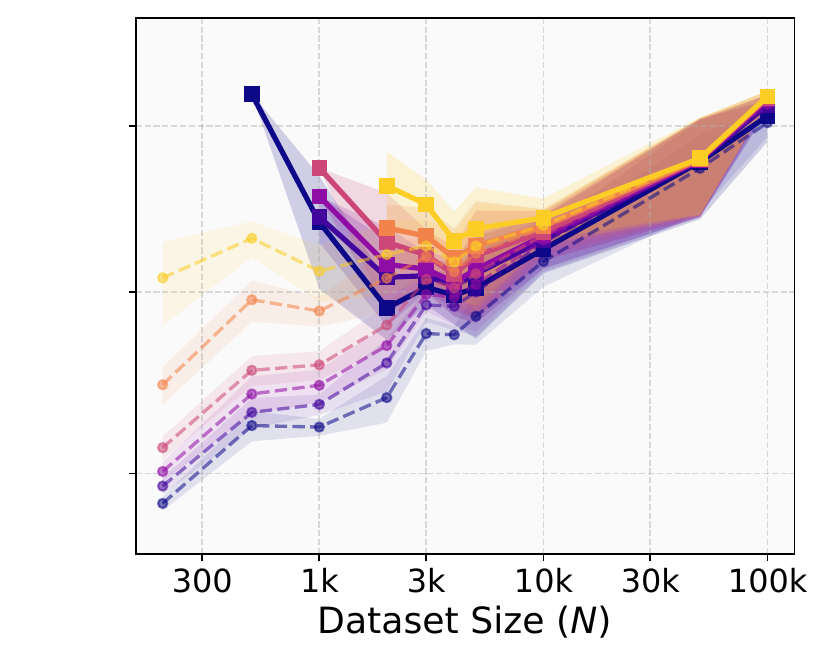}
        \caption{Depth 6.}
        \label{fig:figure9d}
    \end{subfigure}
    \caption{Seed variability for the \(L=5\) NW multi-threshold sweeps.
    Curves show the same depth sweep as Figure~\ref{fig:figure8},
    and shaded regions denote one standard deviation across runs with recorded crossings. Although exact crossing times vary, the qualitative pattern remains stable: validation crossings are non-monotonic in \(N\), and training crossings improve near the onset of weak validation competence.}
    \label{fig:figure9}
\end{figure}

\begin{figure}[!htbp]
    \centering
    \includegraphics[width=\linewidth]{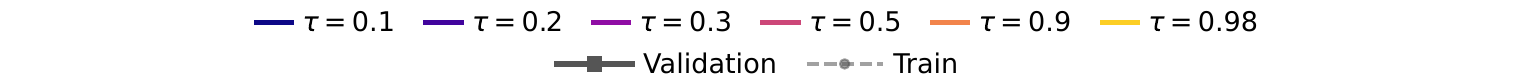}\par\vspace{2pt}
    \begin{subfigure}[t]{0.49\linewidth}
        \centering
        \includegraphics[width=\linewidth]{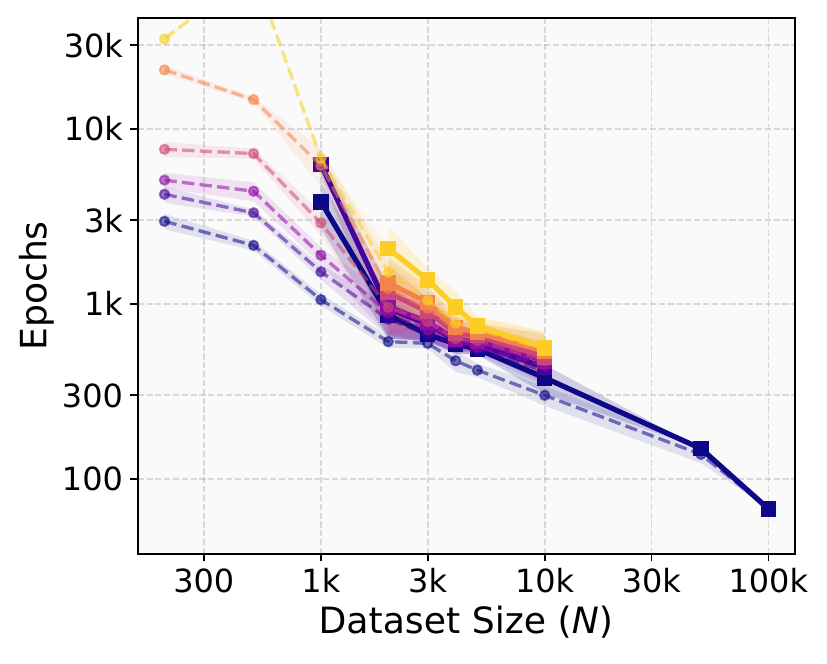}
        \caption{Depth 3.}
        \label{fig:figure10a}
    \end{subfigure}
    \hfill
    \begin{subfigure}[t]{0.49\linewidth}
        \centering
        \includegraphics[width=\linewidth]{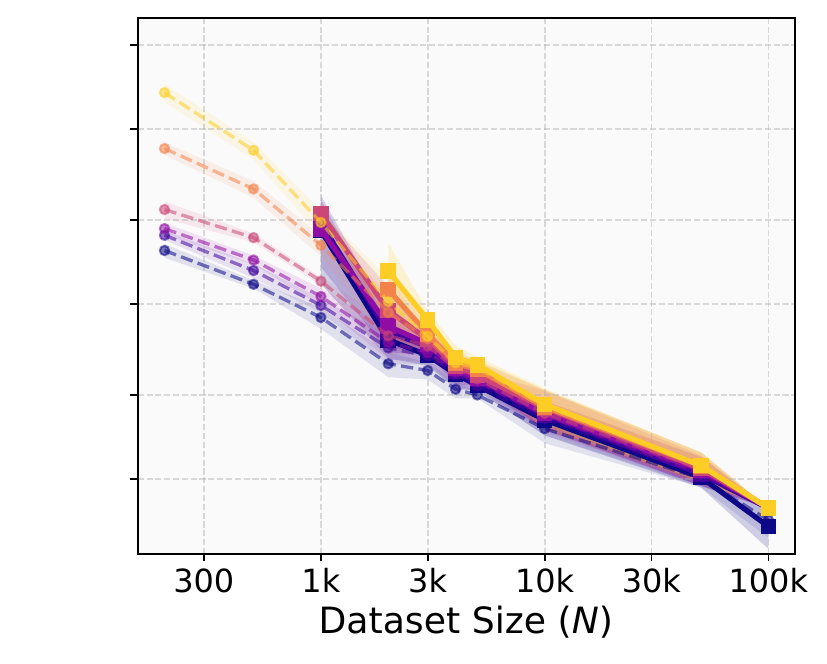}
        \caption{Depth 4.}
        \label{fig:figure10b}
    \end{subfigure}
    \par\vspace{4pt}
    \begin{subfigure}[t]{0.49\linewidth}
        \centering
        \includegraphics[width=\linewidth]{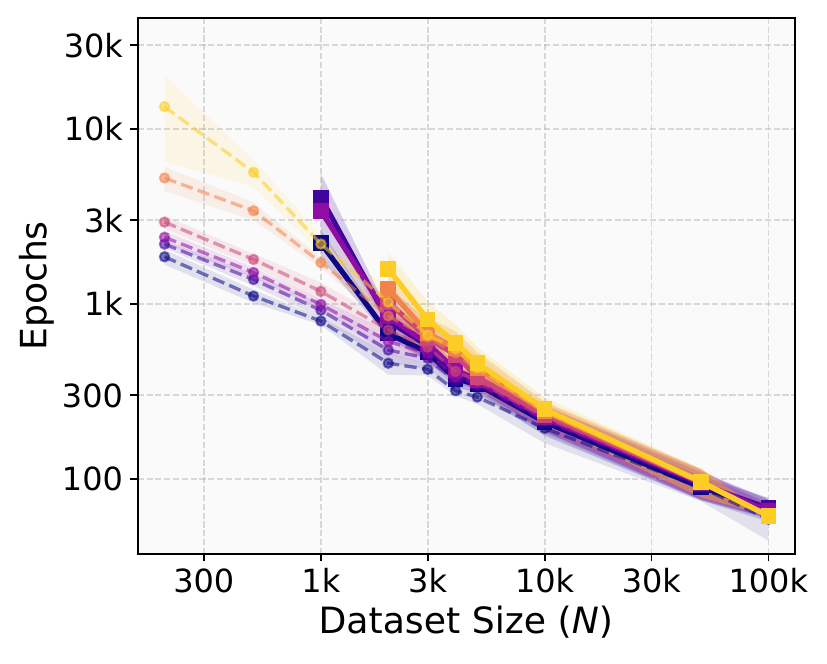}
        \caption{Depth 5.}
        \label{fig:figure10c}
    \end{subfigure}
    \hfill
    \begin{subfigure}[t]{0.49\linewidth}
        \centering
        \includegraphics[width=\linewidth]{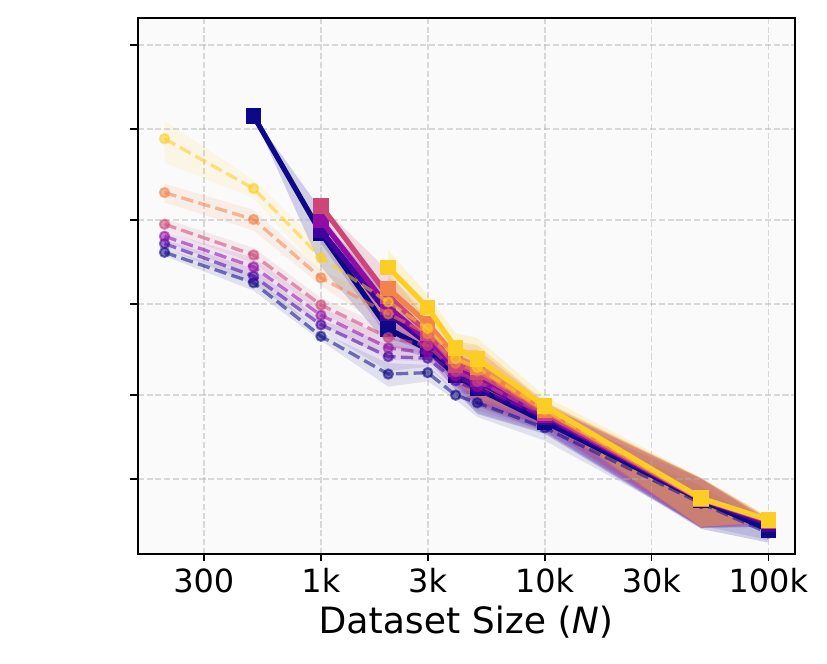}
        \caption{Depth 6.}
        \label{fig:figure10d}
    \end{subfigure}
    \caption{Epoch-normalized convergence for the \(L=5\) NW depth sweep. The same threshold crossings are measured in epochs rather than optimizer updates. Epoch normalization emphasizes data exposure and recovers the more conventional view in which larger datasets require fewer passes over the data. The main-text sweet spot therefore concerns updates, not passes over the data.}
    \label{fig:figure10}
\end{figure}

\FloatBarrier
\section{Robustness Across Sequence Length} \label{app:length}

\begin{figure}[!htbp]
    \centering
    \includegraphics[width=\linewidth]{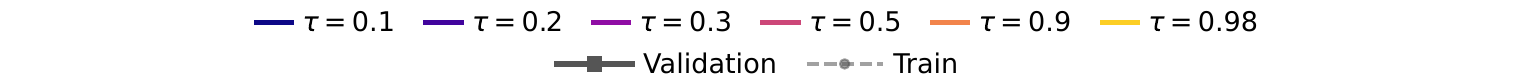}\par\vspace{2pt}
    \begin{subfigure}[t]{0.49\linewidth}
        \centering
        \includegraphics[width=\linewidth]{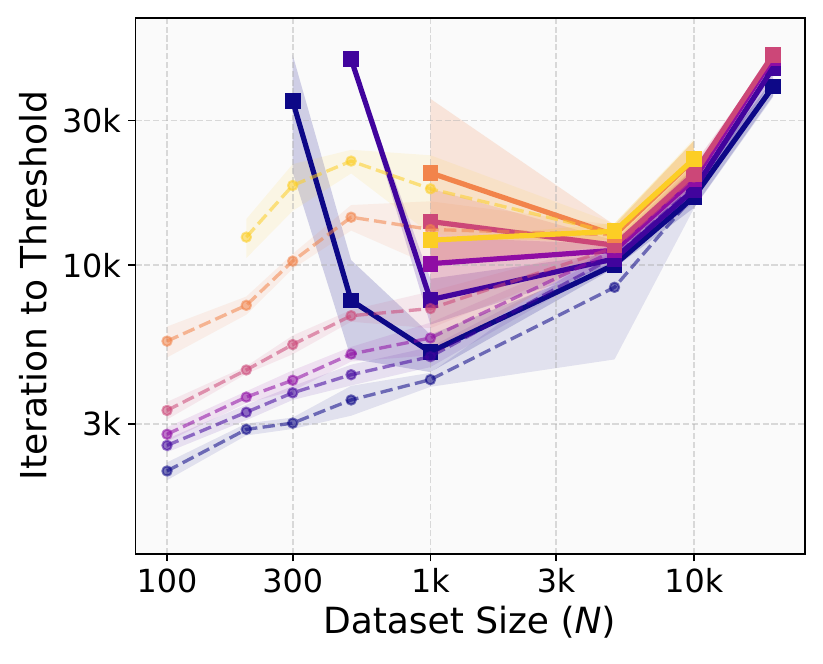}
        \caption{Depth 3.}
        \label{fig:figure11a}
    \end{subfigure}
    \hfill
    \begin{subfigure}[t]{0.49\linewidth}
        \centering
        \includegraphics[width=\linewidth]{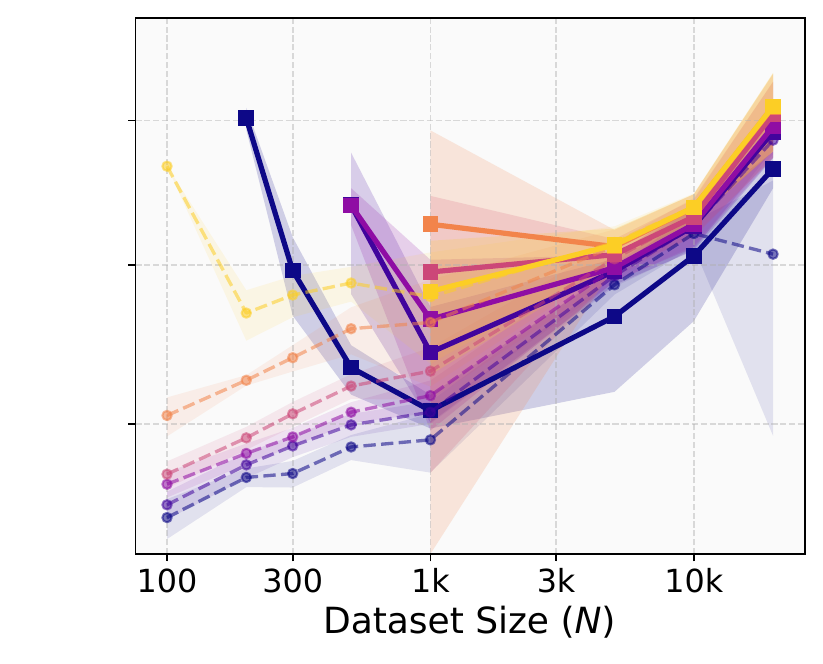}
        \caption{Depth 4.}
        \label{fig:figure11b}
    \end{subfigure}
    \par\vspace{4pt}
    \begin{subfigure}[t]{0.49\linewidth}
        \centering
        \includegraphics[width=\linewidth]{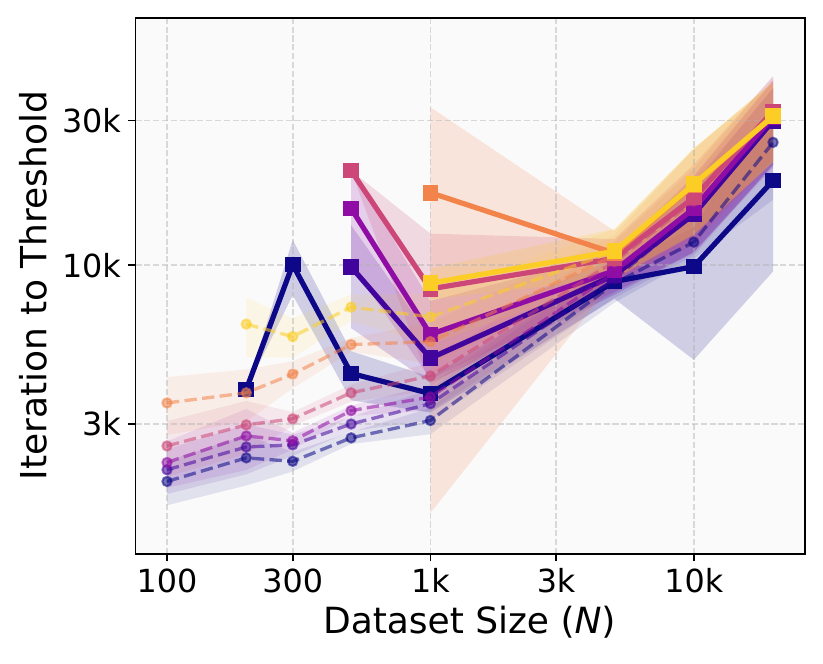}
        \caption{Depth 5.}
        \label{fig:figure11c}
    \end{subfigure}
    \hfill
    \begin{subfigure}[t]{0.49\linewidth}
        \centering
        \includegraphics[width=\linewidth]{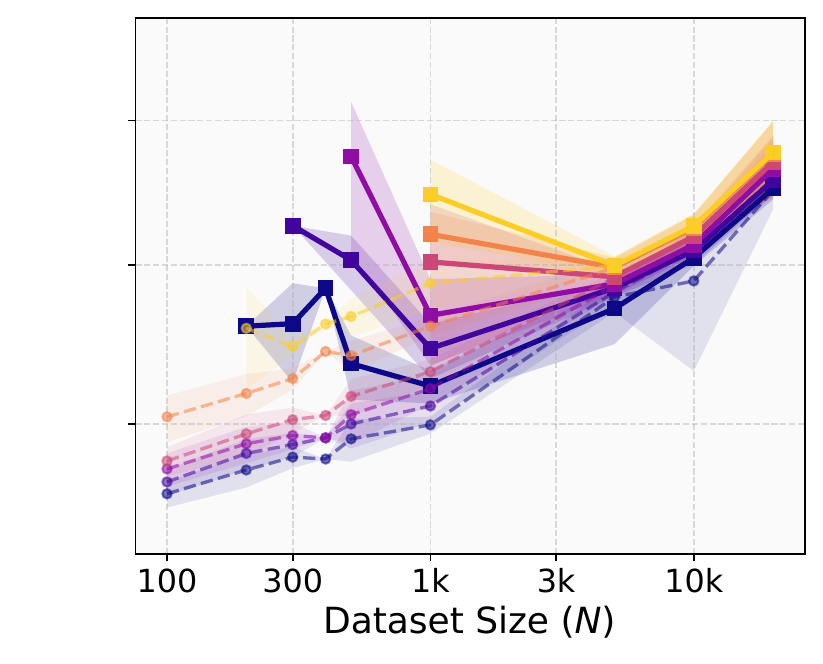}
        \caption{Depth 6.}
        \label{fig:figure11d}
    \end{subfigure}
    \caption{Seed variability for the \(L=4\) NW sweeps. Shaded regions denote one standard deviation across runs with recorded crossings. Exact threshold-crossing times vary, but the intermediate-data regime again shows overlap between weak validation competence and faster training-threshold crossings.}
    \label{fig:figure11}
\end{figure}

\begin{figure}[!htbp]
    \centering
    \includegraphics[width=\linewidth]{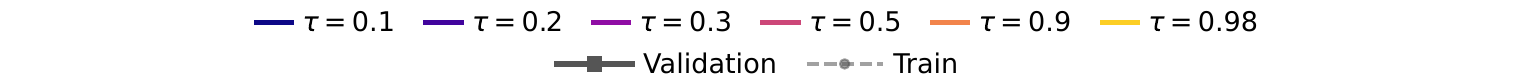}\par\vspace{2pt}
    \begin{subfigure}[t]{0.49\linewidth}
        \centering
        \includegraphics[width=\linewidth]{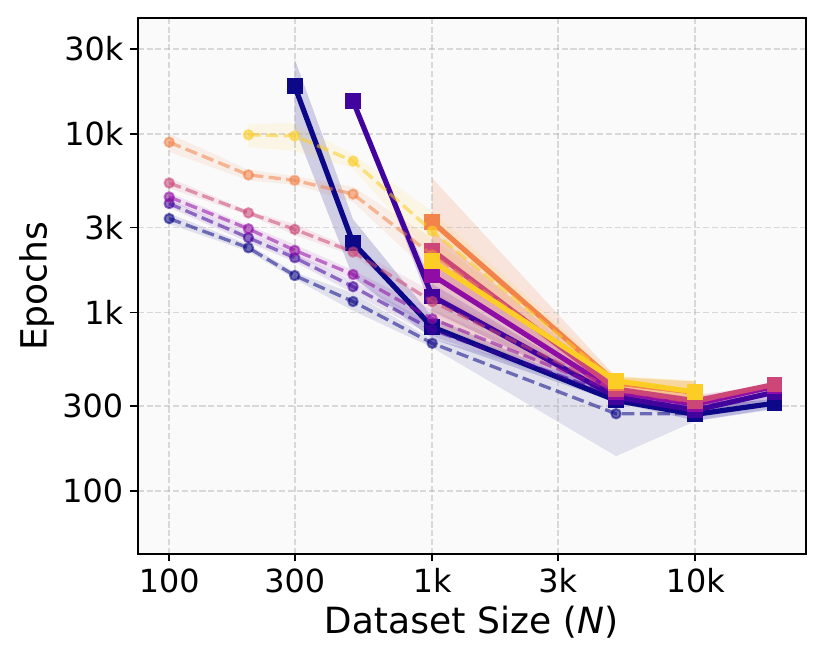}
        \caption{Depth 3.}
        \label{fig:figure12a}
    \end{subfigure}
    \hfill
    \begin{subfigure}[t]{0.49\linewidth}
        \centering
        \includegraphics[width=\linewidth]{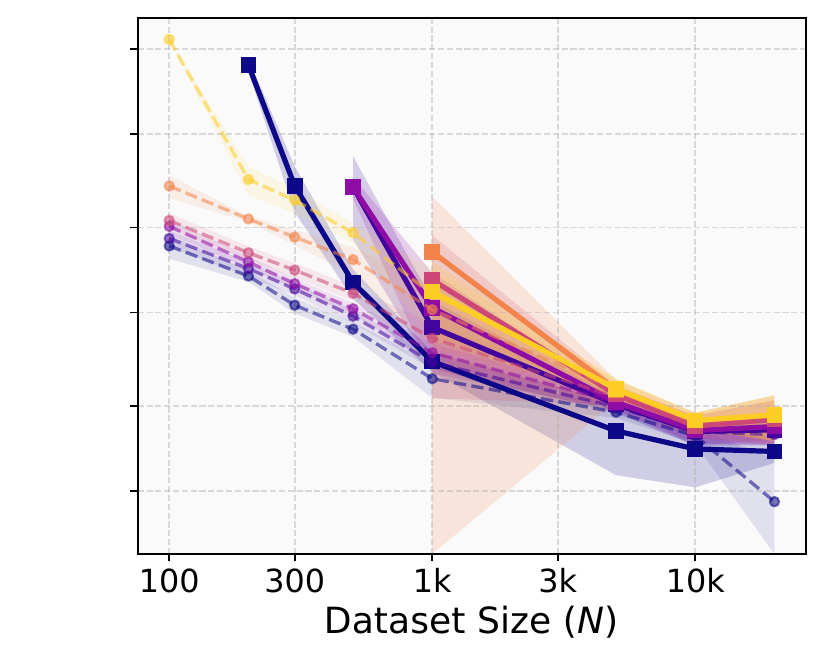}
        \caption{Depth 4.}
        \label{fig:figure12b}
    \end{subfigure}
    \par\vspace{4pt}
    \begin{subfigure}[t]{0.49\linewidth}
        \centering
        \includegraphics[width=\linewidth]{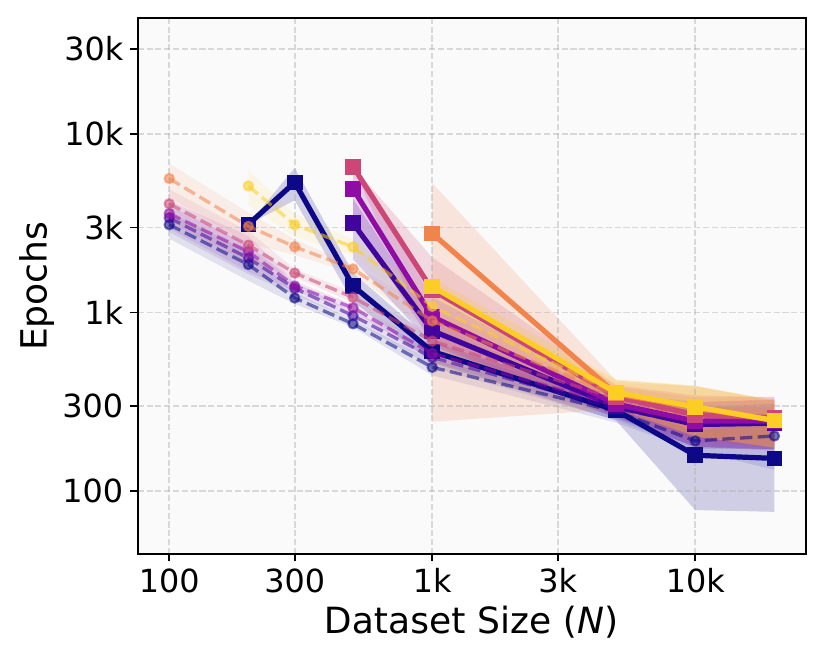}
        \caption{Depth 5.}
        \label{fig:figure12c}
    \end{subfigure}
    \hfill
    \begin{subfigure}[t]{0.49\linewidth}
        \centering
        \includegraphics[width=\linewidth]{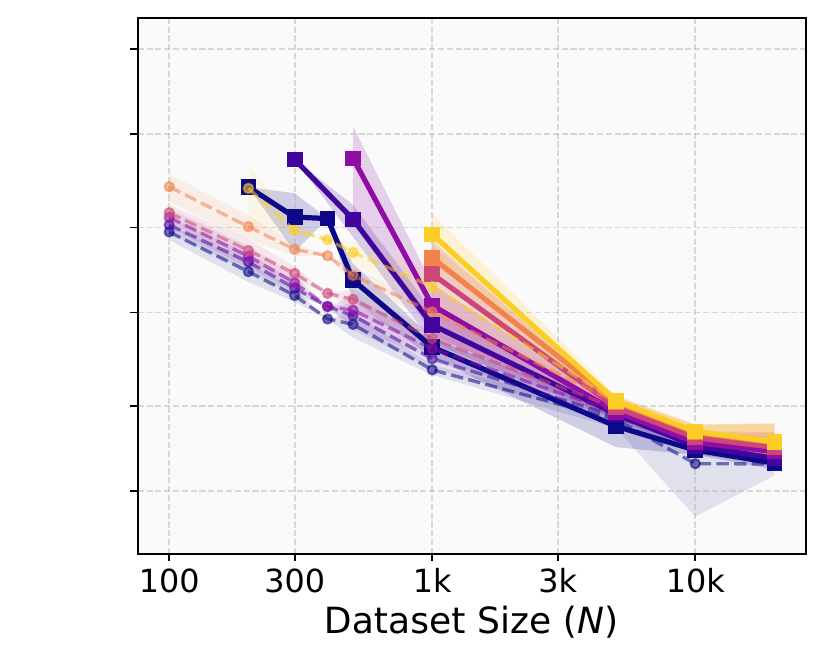}
        \caption{Depth 6.}
        \label{fig:figure12d}
    \end{subfigure}
    \caption{Epoch-normalized convergence for the \(L=4\) NW task. We report epochs to threshold for the same \(L=4\) configurations as Figure~\ref{fig:figure11}. As in the \(L=5\) experiments, epoch normalization emphasizes data exposure rather than parameter-update efficiency and recovers a conventional monotone-data view. This reinforces that the sweet spot studied in the main text is update-based.}
    \label{fig:figure12}
\end{figure}

To test whether the timing pattern is specific to the \(L=5\) Needleman--Wunsch task, we repeat the same analysis at sequence length \(L=4\). 
The full task universe is smaller, but the qualitative structure is similar: validation convergence is non-monotonic in \(N\), and the strict training crossing \(I_T(0.98;N)\) can decrease near the onset of weak validation competence.
Figures~\ref{fig:figure11} and~\ref{fig:figure12} provide seed-variance and epoch-normalized views of the same experiments.

\clearpage
\section{Random-Suffix Experiments} \label{app:suffix}

\begin{figure}[!htbp]
    \centering
    \includegraphics[width=\linewidth]{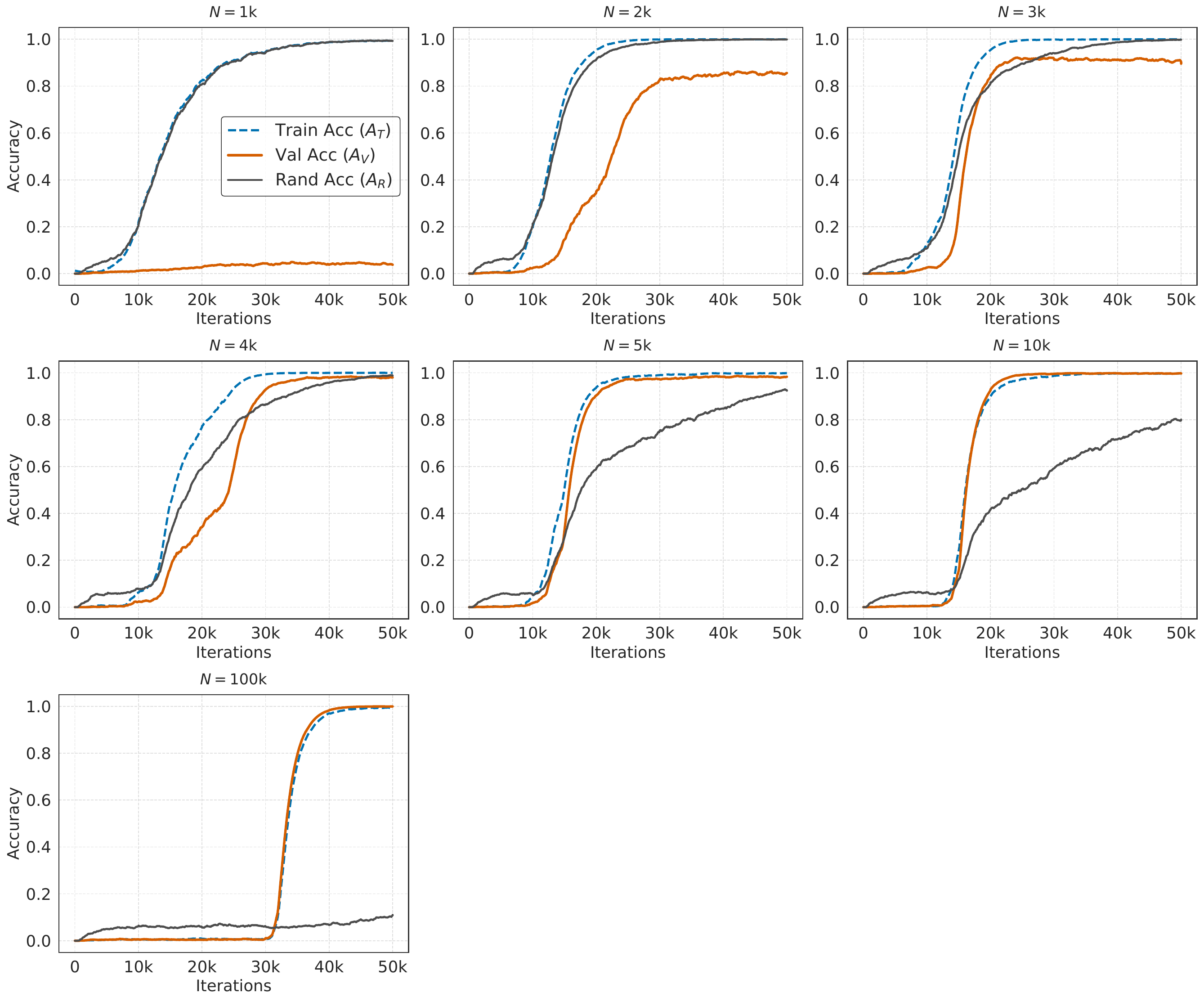}
    \caption{
    Full random-suffix component trajectories. Each NW target is augmented with a four-bit random suffix. We track training accuracy on the structured NW matrix \(A_T\), validation accuracy on the NW matrix \(A_V\), and training accuracy on the random suffix \(A_R\). At small \(N\), \(A_T\) and \(A_R\) rise together while \(A_V\) remains near zero. At larger \(N\), the structured NW component reaches high training accuracy before the much shorter random suffix, matching Figure~\ref{fig:figure3}.
    }
    \label{fig:figure13}
\end{figure}

\begin{figure}[!htbp]
    \centering
    \includegraphics[width=\linewidth]{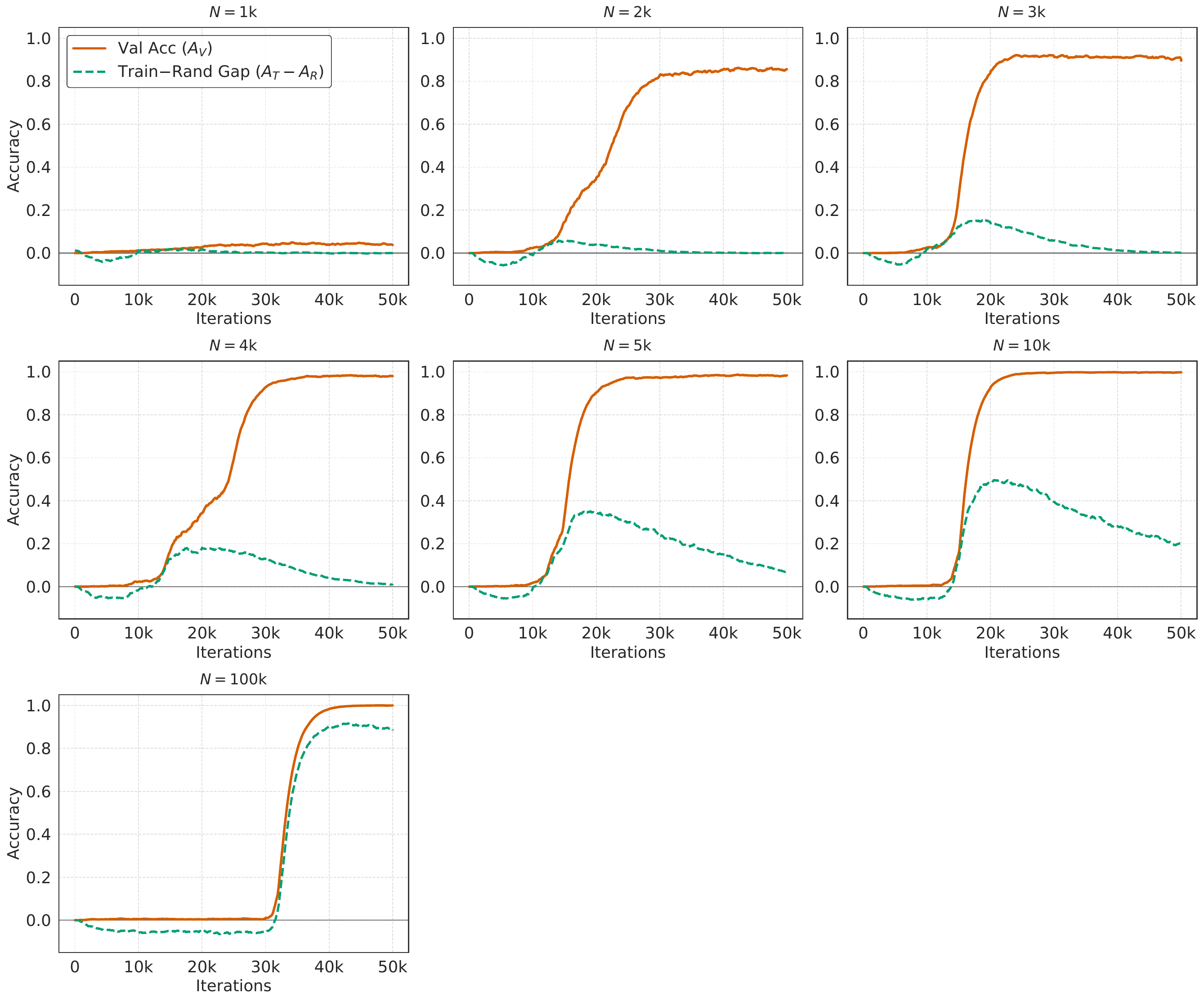}
    \caption{
    Full dataset-size sweep for the train--random gap. For each \(N\), we compare NW validation accuracy \(A_V\) with \(G=A_T-A_R\), where \(A_T\) is training accuracy on the NW matrix and \(A_R\) is training accuracy on the random suffix. The representative cases in Figure~\ref{fig:figure3} reflect the full sweep. At small \(N\), the gap remains small and validation remains low. At larger \(N\), both rise together.
    }
    \label{fig:figure14}
\end{figure}

We extend the main-text random-suffix views to the full dataset-size sweep. Figures~\ref{fig:figure13} and~\ref{fig:figure14} show component-wise trajectories and the train--random gap, respectively.
The suffix provides a separate reference for fitting randomized associations; its accuracy is not a calibrated estimate of residual fitting in the NW component. The correspondence between $A_T-A_R$ and $A_V$ is interpreted locally near acquisition onset.

\FloatBarrier
\section{Task-Scope Controls}
\label{sec:task-specificity}
\label{app:task-scope}

\begin{wrapfigure}{r}{0.43\textwidth}
\vspace{-0.5\baselineskip}
\centering
    \includegraphics[width=\linewidth]{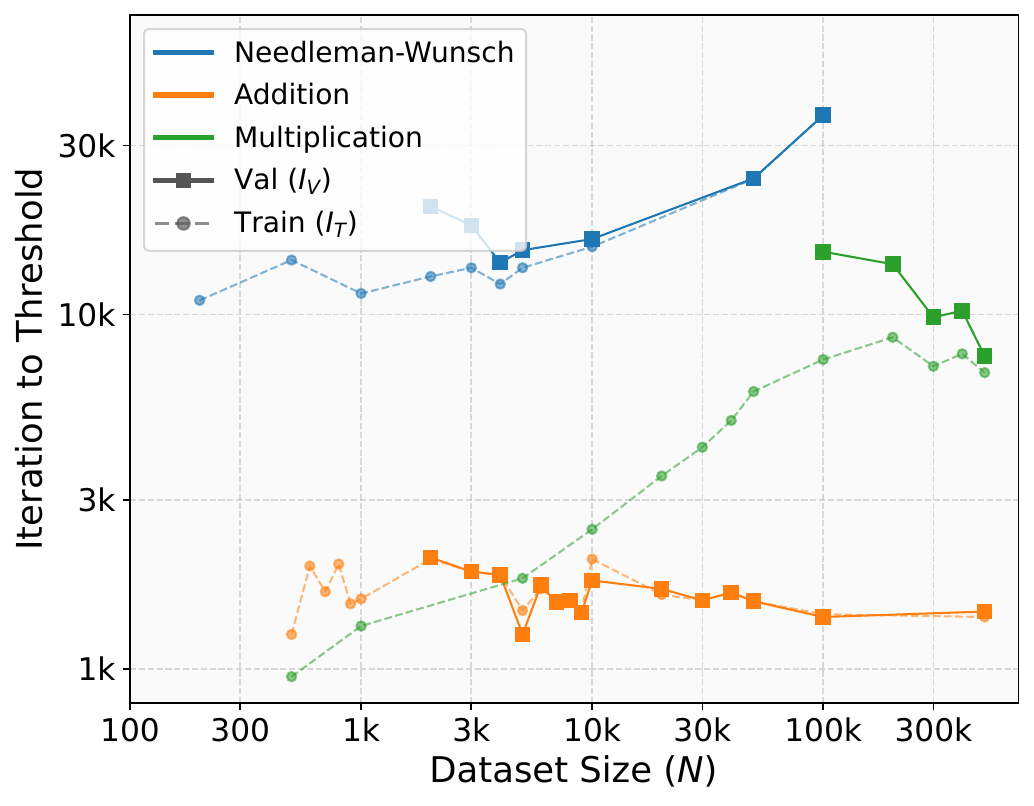}
    \caption{\small
    Task-scope controls at \(\tau=0.98\). 
    Within these sweeps, a sweet spot is observed for NW but not for the direct arithmetic baselines.
    }
    \label{fig:figure15}
\end{wrapfigure}

Figure~\ref{fig:figure15} compares three tasks under exact-match evaluation. Addition has similar training and validation crossing times. Multiplication improves as data grows over much of the tested range, whereas NW has an intermediate-data sweet spot. The task contrasts alone do not isolate the effects of output length or computation. The ECA comparison in Figure~\ref{fig:figure4} holds output length and recurrence structure fixed and still produces different learning regimes.
\looseness=-1\par

The additional edit-distance and scratchpad-multiplication results extend the slowdown beyond NW, and partial NW shows that predicting the full matrix is not required (Appendix~\ref{app:additional-tasks}). The random-suffix probe further shows that a short unstructured target can take longer to fit than a larger rule-generated target. Learning time therefore depends on the predictive signals and example-specific associations available to the model, rather than on output length alone. Rule learning and example-specific fitting can both contribute to accurate prediction, consistent with work on information not captured by shared patterns~\citep{feldman2020neural}.

\section{ECA Experimental and Measurement Details} \label{app:eca-details}

The ECA experiments use rows of length $L=20$ and predict $T=8$ updates from the initial row. The decoder-only Transformer has six layers, six attention heads, and hidden width 384. AdamW uses weight decay $0.1$, with a cosine learning-rate schedule from $10^{-3}$ to $10^{-4}$. The effective batch is 320 complete sequences, distinct from the NW batch size. Within the ECA comparison, all rules use the same initial-row identities, nested dataset sizes, and training settings. Each dataset size crosses three dataset draws with three model initializations.

Exact-trajectory accuracy requires all eight output rows and formatting tokens to be correct in free generation. Validation uses a fixed set separate from training. The plotted training accuracy is measured on a fixed training-dataset diagnostic subset, not the entire training dataset.

For Figure~\ref{fig:figure4}, we identify the highest displayed threshold reached at each rule, size, and split.
The same runs reaching that threshold contribute to the crossing-time means at every lower displayed threshold.
Training and validation are aggregated separately.
Consequently, lower-threshold means condition on reaching the higher threshold, and the contributing cohorts can differ across dataset sizes and splits. 
Unreached thresholds are censored, rather than assigned a finite crossing time.
For Figure~\ref{fig:figure6}, within each training condition, runs reaching validation accuracy $0.98$ contribute to both training and validation means. This pairs the two splits within a condition. These cohort rules differ from the threshold-specific aggregation used for NW.

The population signals in Figure~\ref{fig:figure5} are computed exactly from the eight equally likely three-parent configurations at the first transition. The signal profiles in Figure~\ref{fig:figure6} instead enumerate the initial-row distribution restricted to 13--15 ones and calculate the resulting parent--target distribution separately at each transition. Biased sampling changes the training distribution, while validation remains uniform. 
The intervention preserves the rule and output format but changes multiple properties of the input distribution, so it does not isolate a single $G_k$.

\paragraph{Empirical signals.} For Figure~\ref{fig:figure7}(a), we compute $\widehat G_k=G_k(\widehat P_N)$ from the observed parent--target frequencies in each sampled training dataset. The empirical signals are averaged over the eight transitions within each dataset. Lines and bands then show the median and range over three dataset draws. The plotted values describe prediction under the empirical distribution, whose accidental dependencies need not persist under the uniform population distribution.

\paragraph{Matched Random and Balanced datasets.} For length-20 cyclic rows, Rule~90 has a four-element kernel, so four distinct initial rows produce the same next row. For each sampled first target row $y$, we also use its bitwise complement $\bar y$. Each block contains two inputs mapping to $y$ and two mapping to $\bar y$, so each target trajectory appears twice in both conditions. Balanced selects each input together with its bitwise complement. At every position in the first transition, the construction makes each single-parent--target joint distribution uniform, giving $G_0=G_1=0$. Random instead selects two distinct preimages uniformly for each target, preserving every target trajectory and its multiplicity without enforcing zero single-parent dependence. All initial states are unique. The model and training settings are the same in both conditions.

Both conditions match output frequencies, all subsequent target rows, and trajectory multiplicities. Balanced also changes input marginals and the Hamming distances between inputs sharing a target. The intervention therefore tests the relevance of empirical lower-order signals without isolating $G_1$ from all other input properties. Figure~\ref{fig:figure7}(b) includes all nine training-threshold crossings in each condition. Balancing applies only at the first transition.

\FloatBarrier
\section{Additional Algorithmic Tasks and Partial Outputs}
\label{app:additional-tasks}

\begin{figure}[!htbp]
    \centering
    \includegraphics[width=\linewidth]{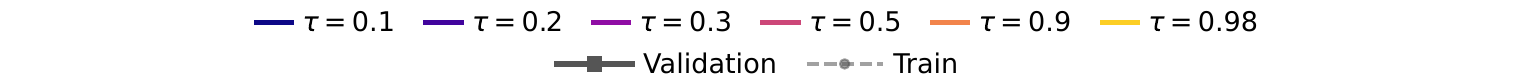}\par\vspace{2pt}
    \begin{subfigure}[t]{0.49\linewidth}
        \centering
        \includegraphics[width=\linewidth]{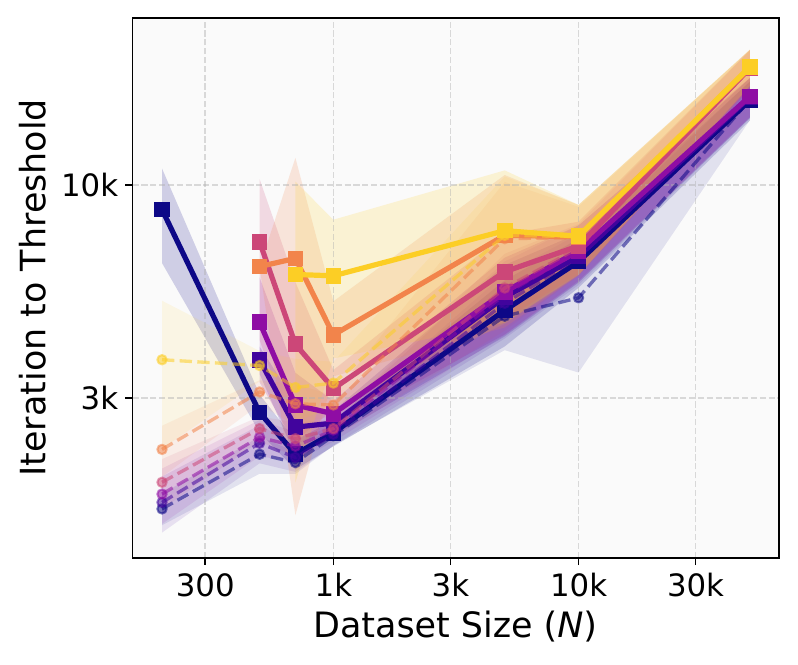}
        \caption{Edit-distance table generation.}
        \label{fig:figure16a}
    \end{subfigure}
    \hfill
    \begin{subfigure}[t]{0.49\linewidth}
        \centering
        \includegraphics[width=\linewidth]{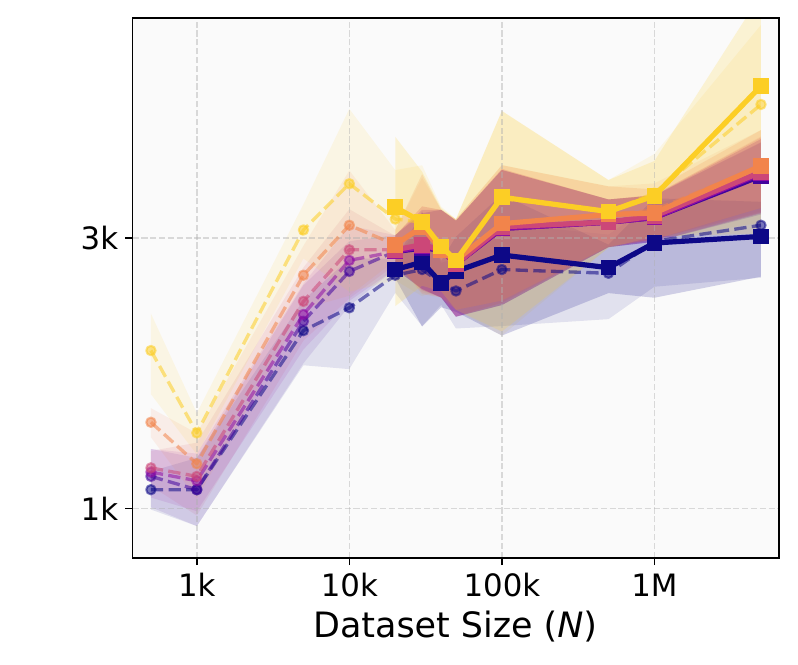}
        \caption{Scratchpad multiplication.}
        \label{fig:figure16b}
    \end{subfigure}
    \caption{Slowdown beyond NW. \textbf{(a)}: edit-distance table generation exhibits an intermediate-data validation optimum. \textbf{(b)}: scratchpad multiplication predicts partial products and their sum, and validation convergence slows at larger dataset sizes. Colors denote exact-output thresholds. Solid squares show validation and dashed circles show training crossings. Lines and shading show means and one standard deviation across recorded crossings. Scratchpad validation at $N=10$k is omitted.}
    \label{fig:figure16}
\end{figure}

We extend the dataset-size sweep to edit-distance table generation and scratchpad multiplication. Edit distance maps two length-four strings over a four-symbol alphabet to the complete Levenshtein cost table, using unit insertion, deletion, and substitution costs~\citep{wagner1974string}. Scratchpad multiplication takes two four-digit integers and generates four place-value partial products followed by their final sum. Edit distance exhibits an intermediate-data validation optimum, and scratchpad multiplication requires more updates at larger dataset sizes (Figure~\ref{fig:figure16}).

Both task sweeps use five runs per dataset size.
For each split and threshold, means and standard deviations use only runs with a recorded crossing, so these summaries condition on attainment within the budget.
The highest displayed threshold, $0.98$, uses the recorded $0.9875$ crossing.
The cutoffs agree for exact match on 80 evaluation examples but need not agree for cell accuracy.
Scratchpad validation at $N=10$k is omitted from Figure~\ref{fig:figure16}; the corresponding measurements are retained in the archived data.

\begin{figure}[!htbp]
    \centering
    \includegraphics[width=0.49\linewidth]{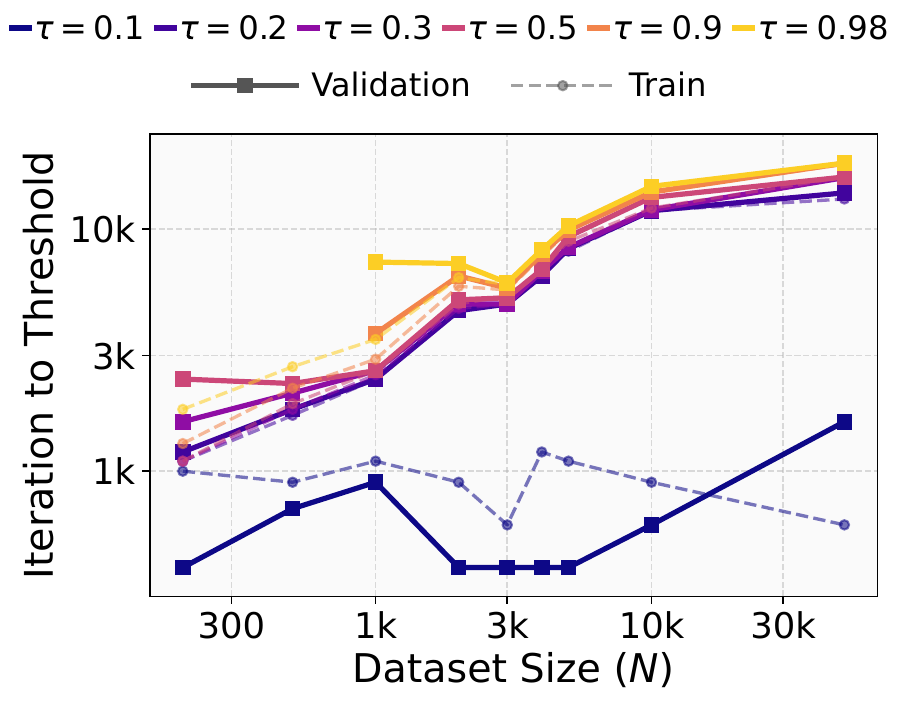}
    \caption{Partial NW predicts only nine cells from the first interior row and column of the NW matrix. Validation crossing time still has an intermediate-data optimum. Colors denote exact-output thresholds, with solid-square validation and dashed-circle training curves. Each curve shows the recorded crossing time for an individual run at each dataset size.}
    \label{fig:figure17}
\end{figure}

The partial-NW experiment predicts nine cells of the length-five NW matrix, comprising its first interior row and the remaining cells of its first interior column. Each dataset size has one recorded run. Figure~\ref{fig:figure17} retains an intermediate-data validation optimum, showing that the observed slowdown can also occur without generating the full matrix. The result complements the ECA comparison, which controls output length and recurrence structure across rules.

\FloatBarrier
\section{Additional NW Robustness Results} \label{app:additional-robustness}

We extend the NW checks to an eight-layer model, cell accuracy, effective batch sizes 80 and 320, and weight decay $0$, $0.01$, and $0.03$. The reference uses six layers, effective batch 160, and weight decay $0.1$. Depth changes follow the model scaling in Appendix~\ref{app:training_details}, including the corresponding attention-head count and hidden width. The remaining comparisons change the indicated setting while keeping the reference architecture and other training settings fixed.

\begin{figure}[!htbp]
    \centering
    \begin{subfigure}[t]{0.49\linewidth}
        \centering
        \includegraphics[width=\linewidth]{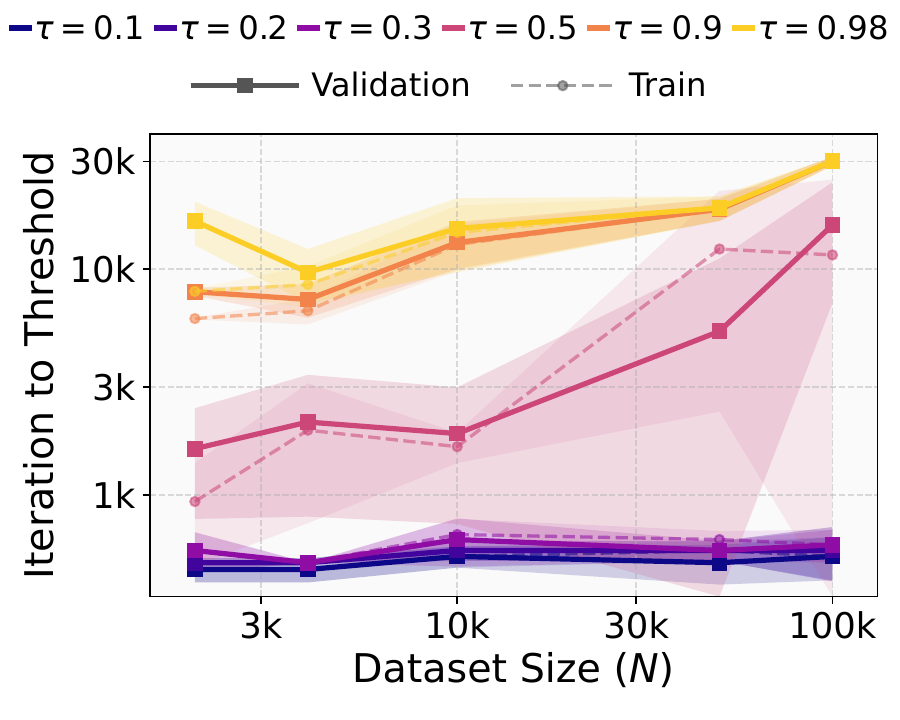}
        \caption{Eight-layer model, exact matrix accuracy.}
        \label{fig:figure18a}
    \end{subfigure}
    \hfill
    \begin{subfigure}[t]{0.49\linewidth}
        \centering
        \includegraphics[width=\linewidth]{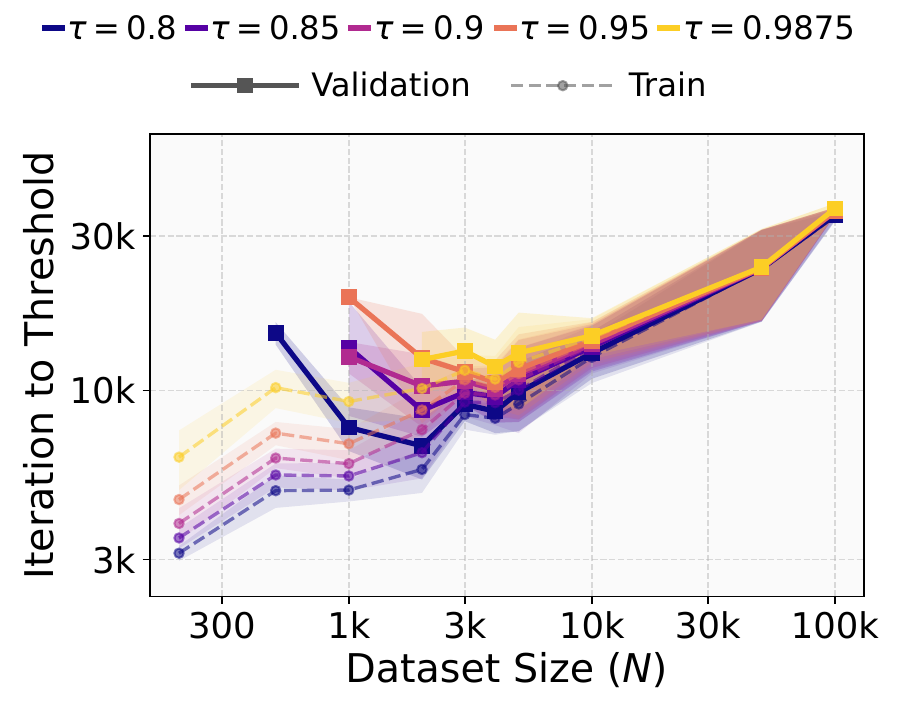}
        \caption{Six-layer model, average cell accuracy.}
        \label{fig:figure18b}
    \end{subfigure}
    \caption{NW capacity and measurement checks. An intermediate-data validation optimum remains visible with eight layers and when accuracy is averaged over matrix cells. Each panel has its own threshold legend because cell accuracy uses higher cutoffs. Solid squares show validation and dashed circles show training crossings, with shading showing one standard deviation.}
    \label{fig:figure18}
\end{figure}

\begin{figure}[!htbp]
    \centering
    \includegraphics[width=\linewidth]{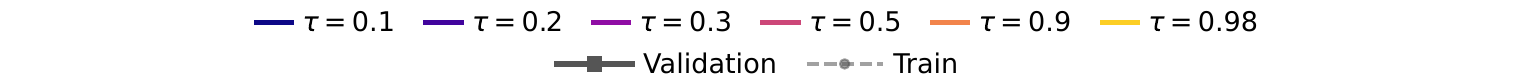}\par\vspace{2pt}
    \begin{subfigure}[t]{0.36663503\linewidth}
        \centering
        \includegraphics[width=\linewidth]{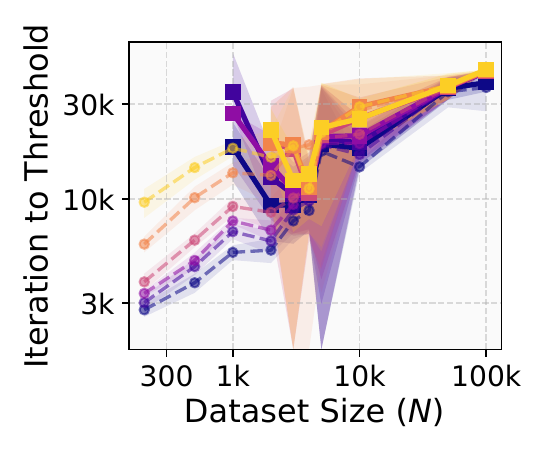}
        \caption{Effective batch 80.}
        \label{fig:figure19a}
    \end{subfigure}%
    \hspace{6pt}%
    \begin{subfigure}[t]{0.30089966\linewidth}
        \centering
        \includegraphics[width=\linewidth]{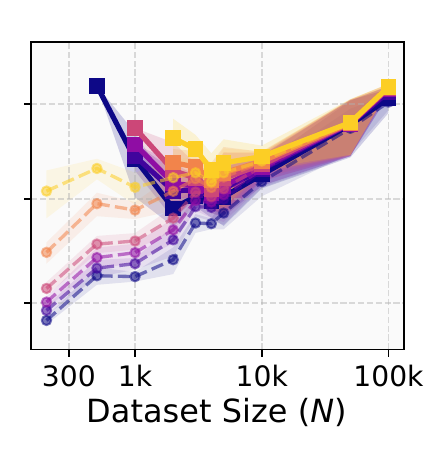}
        \caption{Effective batch 160.}
        \label{fig:figure19b}
    \end{subfigure}%
    \hspace{6pt}%
    \begin{subfigure}[t]{0.30089966\linewidth}
        \centering
        \includegraphics[width=\linewidth]{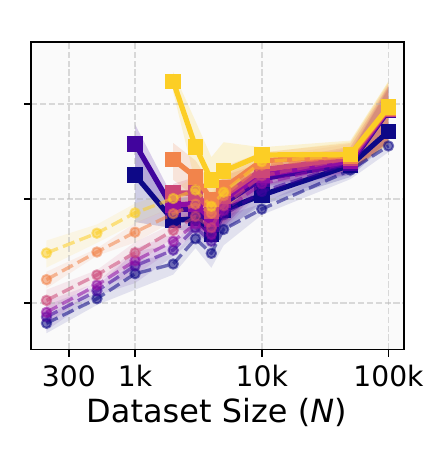}
        \caption{Effective batch 320.}
        \label{fig:figure19c}
    \end{subfigure}
    \caption{NW effective-batch comparison. Panels vary the number of examples per optimizer update across 80, 160, and 320, including gradient accumulation. The six-layer architecture and weight decay $0.1$ are fixed. The intermediate-data optimum shifts, and validation still takes more updates at larger dataset sizes. Colors denote exact-matrix thresholds. Solid-square validation and dashed-circle training curves show means, with one-standard-deviation shading.}
    \label{fig:figure19}
\end{figure}

The eight-layer sweep uses three runs per dataset size.
The cell-accuracy and optimization sweeps use five, while the reference has four recorded runs at $N=10$k and five elsewhere.
At each threshold, training and validation means are computed separately over runs with a recorded crossing. 
Shading shows one standard deviation of those crossing times. The number of contributing runs can vary across sizes, splits, and thresholds.

Across the batch-size and positive-weight-decay comparisons, validation curves rise again at larger dataset sizes, although the fastest size shifts. With zero weight decay, high-threshold crossings are recorded for fewer runs, especially at smaller sizes. The curves support slowdown under several optimization settings, with stronger evidence where the threshold is attained more consistently.

\begin{figure}[!htbp]
    \centering
    \includegraphics[width=\linewidth]{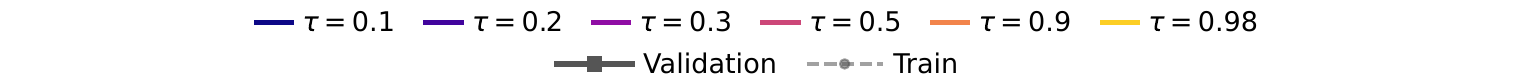}\par\vspace{2pt}
    \begin{subfigure}[t]{0.49\linewidth}
        \centering
        \includegraphics[width=\linewidth]{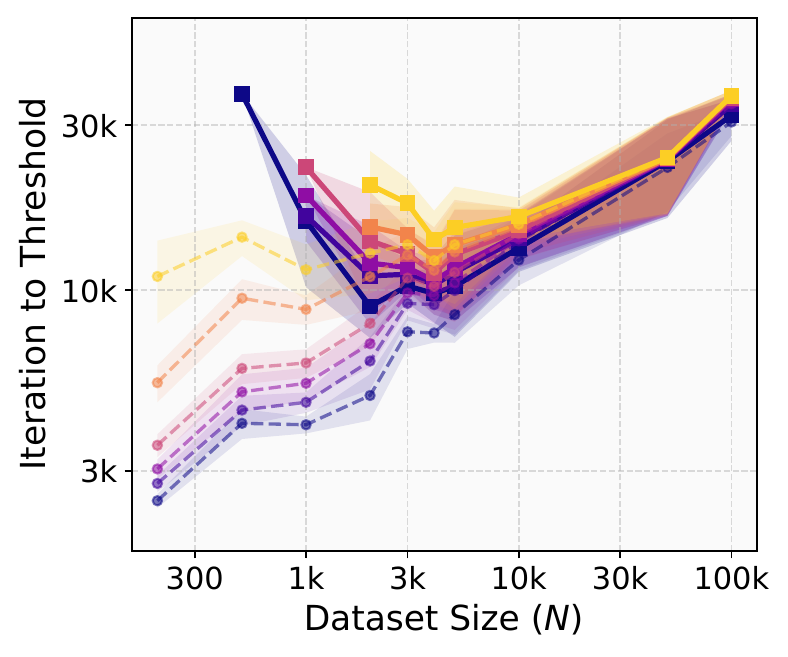}
        \caption{Weight decay $0.1$.}
        \label{fig:figure20a}
    \end{subfigure}
    \hfill
    \begin{subfigure}[t]{0.49\linewidth}
        \centering
        \includegraphics[width=\linewidth]{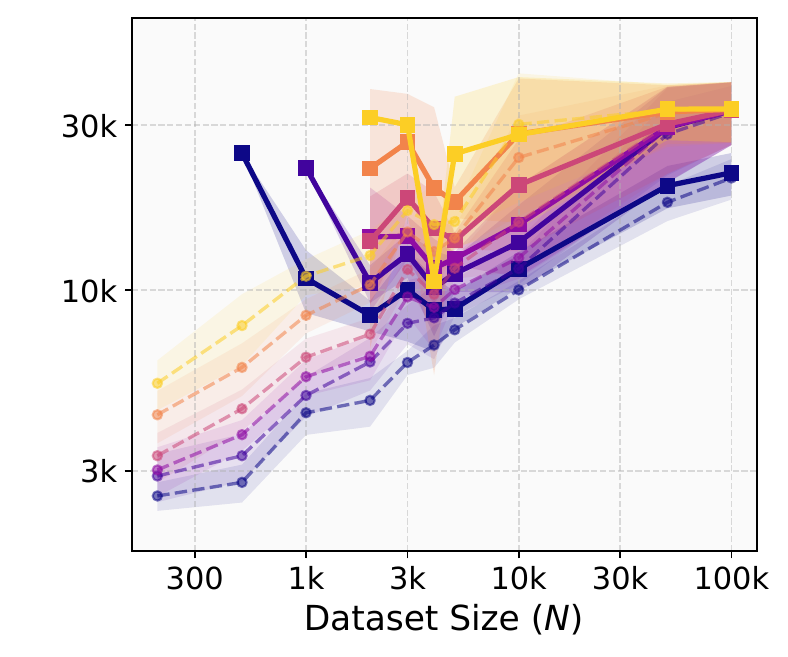}
        \caption{Weight decay $0$.}
        \label{fig:figure20b}
    \end{subfigure}
    \par\vspace{4pt}
    \begin{subfigure}[t]{0.49\linewidth}
        \centering
        \includegraphics[width=\linewidth]{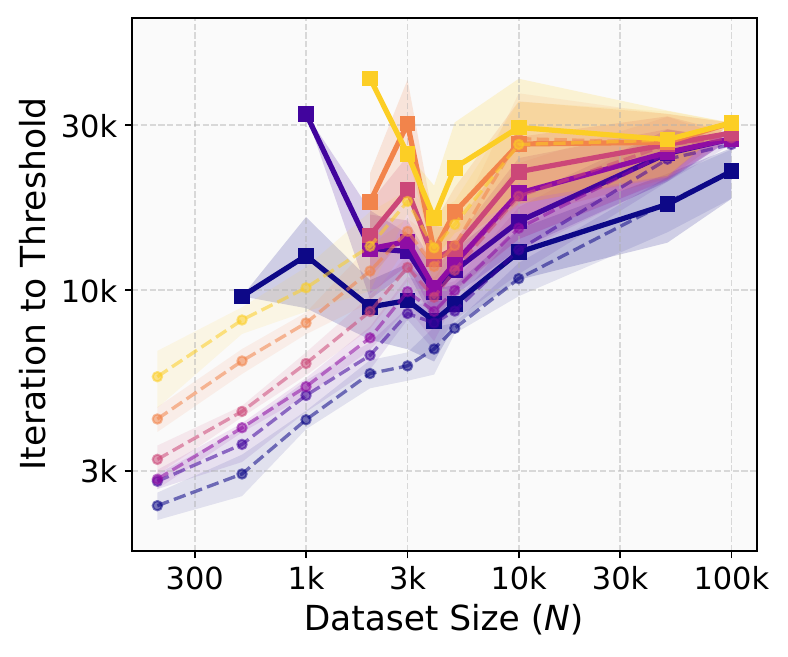}
        \caption{Weight decay $0.01$.}
        \label{fig:figure20c}
    \end{subfigure}
    \hfill
    \begin{subfigure}[t]{0.49\linewidth}
        \centering
        \includegraphics[width=\linewidth]{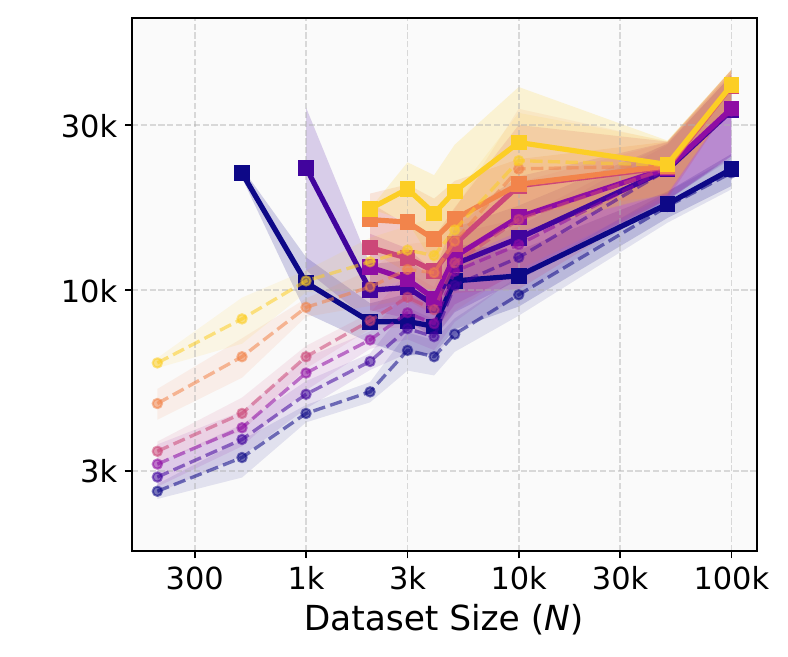}
        \caption{Weight decay $0.03$.}
        \label{fig:figure20d}
    \end{subfigure}
    \caption{NW weight-decay comparison. The six-layer model and effective batch 160 are fixed, and panels vary weight decay across $0.1$, $0$, $0.01$, and $0.03$. Large-data delay appears in several settings, although the fastest size and threshold attainment change. Colors denote exact-matrix thresholds. Solid-square validation and dashed-circle training curves show means and one-standard-deviation shading across recorded crossings. High-threshold means with zero weight decay use fewer runs.}
    \label{fig:figure20}
\end{figure}

\FloatBarrier
\section{Broader impact}
This work is foundational and studies synthetic algorithmic datasets rather than deployed systems. Its main potential benefit is a clearer understanding of memorization--generalization dynamics under controlled conditions. We do not release pretrained models, scraped datasets, personal data, or user-facing systems. We therefore do not anticipate direct societal harms, although the general study of memorization may inform future work on privacy, robustness, and data efficiency.






\end{document}